\documentclass[10pt,journal,compsoc]{IEEEtran}
\ifCLASSOPTIONcompsoc
  \usepackage[nocompress]{cite}
\else
  \usepackage{cite}
\fi
\ifCLASSINFOpdf
\else
\fi

\usepackage{amsmath,amsfonts}
\usepackage{algorithmic}
\usepackage{algorithm}
\usepackage{array}
\usepackage[caption=false,font=normalsize,labelfont=sf,textfont=sf]{subfig}
\usepackage{textcomp}
\usepackage{stfloats}
\usepackage{url}
\usepackage{verbatim}
\usepackage{graphicx}
\usepackage{cite}
\usepackage{diagbox}
\usepackage{subfig}

\usepackage{graphicx}
\usepackage{multirow}

\usepackage{microtype}
\usepackage{graphicx}
\usepackage{booktabs} % for professional tables
\usepackage{amsmath}

\usepackage{dsfont}
\usepackage{bm}
\usepackage{multirow}
\usepackage{bbding}
\usepackage{color}
\usepackage{bbding}
\usepackage{pifont}  % ✔ & ×
\usepackage{bbm}
\usepackage{amsmath}
\usepackage{amssymb}
\usepackage{mathtools}
\usepackage{amsthm}
\usepackage{ragged2e}

\theoremstyle{plain} % Default style
\newtheorem{theorem}{Theorem}
\newtheorem{proposition}[theorem]{Proposition}

\theoremstyle{definition}

\theoremstyle{remark}

\usepackage{hyperref}

\begin{document}
\title{Towards Accurate Post-Training Quantization of Vision Transformers via Error Reduction}

\author{Yunshan Zhong,
You Huang,
Jiawei Hu,
Yuxin Zhang,
Rongrong Ji,~\IEEEmembership{Senior Member, IEEE}
\thanks{Y. Zhong is with the Institute of Artificial Intelligence, Department of Artificial Intelligence, School of Informatics, and Key Laboratory of Multimedia Trusted Perception and Efficient Computing, Ministry of Education of China, Xiamen University, Xiamen 361005, P.R. China.}
\thanks{Y. Huang, J. Hu, Y. Zhang are with the Department of Artificial Intelligence, School of Informatics, and Key Laboratory of Multimedia Trusted Perception and Efficient Computing, Ministry of Education of China, Xiamen University, Xiamen 361005, P.R. China.}
\thanks{R. Ji (Corresponding Author) is with the Institute of Artificial Intelligence, and Key Laboratory of Multimedia Trusted Perception and Efficient Computing, Ministry of Education of China, Xiamen University, Xiamen 361005, P.R. China, and also with the Peng Cheng Laboratory, Shenzhen 518000, P.R. China (e-mail: rrji@xmu.edu.cn).}
}


\markboth{Journal of \LaTeX\ Class Files,~Vol.~14, No.~8, August~2015}%
{Shell \MakeLowercase{\textit{et al.}}: Bare Demo of IEEEtran.cls for Computer Society Journals}



\IEEEtitleabstractindextext{%
\begin{abstract} \justifying{Post-training quantization (PTQ) for vision transformers (ViTs) has received increasing attention from both academic and industrial communities due to its minimal data needs and high time efficiency. However, many current methods fail to account for the complex interactions between quantized weights and activations, resulting in significant quantization errors and suboptimal performance.
This paper presents ERQ, an innovative two-step PTQ method specifically crafted to reduce quantization errors arising from activation and weight quantization sequentially.
The first step, Activation quantization error reduction (Aqer), first applies Reparameterization Initialization aimed at mitigating initial quantization errors in high-variance activations. Then, it further mitigates the errors by formulating a Ridge Regression problem, which updates the weights maintained at full-precision using a closed-form solution.
The second step, Weight quantization error reduction (Wqer), first applies Dual Uniform Quantization to handle weights with numerous outliers, which arise from adjustments made during Reparameterization Initialization, thereby reducing initial weight quantization errors.
Then, it employs an iterative approach to further tackle the errors. In each iteration, it adopts Rounding Refinement that uses an empirically derived, efficient proxy to refine the rounding directions of quantized weights, complemented by a Ridge Regression solver to reduce the errors.
Comprehensive experimental results demonstrate ERQ's superior performance across various ViTs variants and tasks. For example, ERQ surpasses the state-of-the-art GPTQ by a notable 36.81\% in accuracy for W3A4 ViT-S. Our codes are available at \url{https://github.com/zysxmu/ERQ}.}
\end{abstract}

}

\maketitle

\IEEEdisplaynontitleabstractindextext

% \fi
\IEEEpeerreviewmaketitle

\IEEEraisesectionheading{\section{Introduction}\label{sec:introduction}}

\IEEEPARstart{I}{n} the realm of computer vision, vision transformers (ViTs)~\cite{DosovitskiyZ21An} have emerged as the new fundamental backbone models, significantly challenging the convolutional neural networks (CNNs). By leveraging multi-head self-attention (MHSA) mechanism to capture long-range relationships, ViTs exhibit strong and flexible representation capacity, thus resulting in impressive progress in a variety of vision tasks~\cite{liu2021swin,touvron2021training,carion2020end,zhu2020deformable,zheng2021rethinking,arnab2021vivit}. 
However, ViTs' great power comes with considerable complexity. 
The intricate architecture and large number of parameters of ViTs result in high computational and memory demands.
As a result, deploying ViTs in resource-constrained environments such as mobile phones becomes a huge challenge~\cite{tang2022patch,jia2021efficient,hou2022multi,zheng2023less,li2023vit,liu2021post,chen2023diffrate}.

\begin{figure}[t]
\begin{center}
\centerline{\includegraphics[width=1\columnwidth]{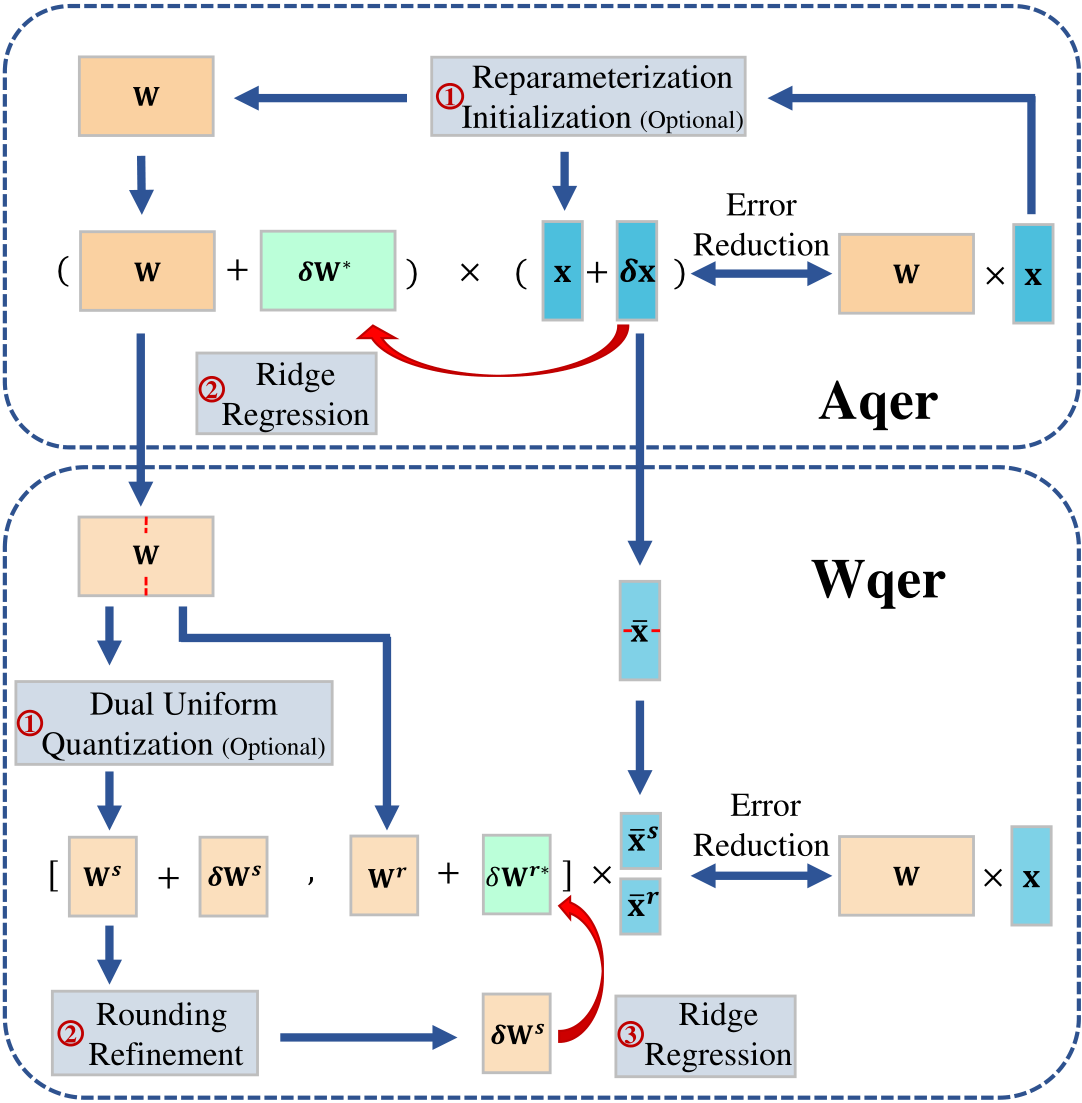}}
\caption{Framework of the proposed ERQ. ERQ consists of two steps to reduce the quantization errors from activation and weight quantization, respectively. The first step, Activation quantization error reduction (Aqer), includes Reparameterization Initialization and Ridge Regression. The second step, Weight quantization error reduction (Wqer), includes Dual Uniform Quantization, Rounding Refinement, and Ridge Regression.}
\label{fig:framework}
\end{center}
\end{figure}

To mitigate this dilemma, the model quantization technique has received its popularity~\cite {whitepaper}. Quantization reduces model complexity by using a low-bit data format to represent the full-precision weights and activations. Consequently, quantization not only facilitates more efficient low-bit computation but also significantly reduces storage requirements, offering a promising solution for deployment in resource-limited devices~\cite{LSQ}. However, quantization is known to introduce notorious errors, which can significantly degrade model performance~\cite{ACIQ}. To mitigate quantization errors, Quantization-aware training (QAT) trains the quantized model with the full original dataset, effectively retaining performance~\cite{li2022q,li2023vit,APoT,gong2019differentiable,Yang2023Oscillation,LSQ,huang2023variation}. However, QAT's reliance on complete training datasets and compute-intensive retraining poses significant limitations~\cite{li2023psaqv2}.
Recently, to address the high demands of QAT, researchers have been gravitating towards post-training quantization (PTQ), which aims to quantize models using only a small calibration dataset and incurs minimal time costs~\cite{li2022patch,lin2022fqvit,liu2023noisyquant,frumkin2023jumping,tang2024easyquant}.

Given the distinct structure of ViTs, such as MHSA and layer normalization (LN), traditional PTQ methods developed for CNNs often fall short when applied to ViTs, resulting in limited performance~\cite{liu2021post}. 
Therefore, researchers have developed specialized PTQ methods for ViTs. For example, the log2 quantizer~\cite{lin2022fqvit,li2023repq} and the twin uniform quantizer~\cite{yuan2022ptq4vit} are introduced for handling long-tail post-Softmax activations. To manage high variant activations, the power-of-two factor is employed~\cite{lin2022fqvit}. For determining unstable scale factors, evolutionary search methods~\cite{frumkin2023jumping} are utilized. 
Despite these advancements, current methods suffer from considerable quantization error due to the overlook of the intricate interplay between weight and activation quantization. As a result, the performance of the quantized model remains unsatisfactory.

In this paper, we introduce \textbf{ERQ}, a novel two-step PTQ method tailored for ViTs, that sequentially mitigates quantization error induced by quantized activations and weights, respectively.
As depicted in Fig.\,\ref{fig:framework}, ERQ consists of two sequential steps: Activation quantization error reduction (Aqer) followed by Weight quantization error reduction (Wqer). 
In Aqer, we focus on quantizing activations while maintaining weights at full-precision to solely mitigate the quantization errors arising from activation quantization.
Initially, specifically for post-LayerNorm activations that exhibit high channel variance~\cite{lin2022fqvit}, we introduce Reparameterization Initialization. This approach leverages the reparameterization technique~\cite{li2023repq}, which adopts the channel-wise quantizer at first and then translates it to the layer-wise equivalence through equivalent change between LN and weights, to set the quantizer for a low initial quantization error.
Subsequently, we further mitigate the errors by formulating a Ridge Regression problem, which offers a closed-form solution. This solution is then applied to update the full-precision weights, effectively reducing the quantization errors.

Following Aqer, we then perform weight quantization and mitigate the induced quantization errors. 
Initially, we identify that weights adjusted via Reparameterization Initialization are prone to extensive outliers, incurring significant initial weight quantization errors under standard uniform quantization. 
To address this, we introduce Dual Uniform Quantization, which separates channels with outliers from those without, assigning individual quantization parameters to each type of channel, for a low initial quantization error. Moreover, an outlier channel selection algorithm with theoretically optimal outlier coverage is introduced to select the outlier channels.
Subsequently, the errors are further mitigated in an iterative quantization-and-correction manner. In each iteration, the first half of the full-precision weights is quantized and the resulting errors are mitigated by first performing Rounding Refinement and then again solving a Ridge Regression problem. The former derives an efficient proxy for output error, which aids in refining the rounding directions of quantized weights to reduce the quantization errors. The latter mitigates the quantization errors by updating the remaining second half of full-precision weights. Such a process continuously performs until all weights are accurately quantized.

The extensive experiments across various ViT variants (ViT, DeiT, and Swin) and tasks (image classification, object detection, instance segmentation, and image super-resolution) show that the proposed ERQ achieves a marked improvement over previous state-of-the-art approaches, highlighting its versatility and robustness. For instance, in the image classification task, ERQ achieves a 36.81\% accuracy gain for the W3A4 ViT-S. In object detection and instance segmentation tasks, ERQ enhances performance by 5.0 AP$^\text{box}$ and 4.8 AP$^\text{mask}$ respectively, when applied to W3A4 Mask R-CNN using a Swin-T backbone. Furthermore, in the image super-resolution task, ERQ increases the PSNR by 0.54 dB on the Urban dataset for W4A4 SwinIR$\times$2.

This work extends from our previous work in \cite{zhongerq}. The six new contributions include: 
(1) We offer a detailed description of the Reparameterization Initialization employed in our method, along with an in-depth discussion of motivation and adequate experimental analysis; 
(2) We propose a new Dual Uniform Quantization method and design a channel selection algorithm for optimal outlier coverage, accompanied by an in-depth discussion of the motivation and comprehensive experimental validation;
(3) We conduct additional quantitative and qualitative experiments on the image super-resolution task to further validate the effectiveness of our approach;
(4) We perform an extensive set of experiments across multiple bit-width settings on object detection and instance segmentation, and add more recent methods in comparison;
(5) We provide comprehensive ablation studies to investigate the effectiveness of each component within ERQ;
(6) We present a detailed illustration of the overall framework of the proposed ERQ, enhancing understanding and applicability.

\section{Related Work}

\subsection{Vision Transformers (ViTs)}

Inspired by the success of transformers in natural language processing, ViTs have emerged as a groundbreaking development in computer vision~\cite{DosovitskiyZ21An}. To address the dependency of ViTs on large datasets, DeiT~\cite{touvron2021training} showcases a teacher-student training approach, offering an efficient data-driven training strategy.
Subsequently, Swin Transformer~\cite{liu2021swin} incorporates a hierarchical structure with a shifted window-based self-attention mechanism, marking further advancements.
Beyond the classification tasks, the applications of ViTs have expanded considerably, including areas such as object detection~\cite{carion2020end,zhu2020deformable}, image segmentation~\cite{chen2021pre,zheng2021rethinking}, low-level image processing~\cite{liang2021swinir}, video classification~\cite{neimark2021video,arnab2021vivit}, and medical image processing~\cite{shamshad2023transformers}, among others.
However, ViTs are accompanied by substantial computational overhead and increased memory requirements, posing challenges for their deployment in resource-constrained environments.
Despite numerous efforts to develop lightweight ViTs, such as MiniVit~\cite{zhang2022minivit}, MobileViT~\cite{mehta2021mobilevit}, and TinyViT \cite{wu2022tinyvit}, the complexity remains a  concern~\cite{chen2023smmix,li2022patch}.

\subsection{Post-training Quantization for ViTs}

Model quantization reduces the numerical precision of weights and activations to decrease computational and storage demands of neural networks~\cite{whitepaper}. However, the process of converting input full-precision data into low-bit data format inevitably introduces notorious quantization errors, which can undesirably degrade model performance. To counteract these errors, a common way is to train the quantized model with the full original training data to accommodate the model with the quantization errors, which is known as quantization-aware training (QAT)~\cite{li2022q,li2023vit,APoT,gong2019differentiable,Yang2023Oscillation,LSQ,huang2023variation}. While QAT effectively maintains model performance, it is vulnerable to the necessity for complete training data and compute-heavy retraining~\cite{li2023psaqv2}.

In contrast to QAT which involves complete training data and compute-heavy retraining, post-training quantization (PTQ) operates on a smaller dataset with a reduced time overhead, harvesting extensive attention~\cite{ACIQ}.
Unfortunately, the unique architectural intricacies of ViTs, such as MHSA and LN, make that conventional PTQ methods, tailored for CNNs, often underperform when applied to ViTs~\cite{li2021brecq,wei2021qdrop,ramachandran2024clamp,li2023repq,lin2022fqvit}.
Consequently, the need for specialized PTQ methods fo ViTs has been recognized, sparking extensive research in this direction. 
Liu \emph{et al}.~\cite{liu2021post} introduce the first PTQ method for ViTs. To maintain the order of softmax scores and adapt various quantization sensitivities of different layers, they respectively introduce a ranking loss and a nuclear norm-based mixed-precision scheme.
FQ-ViT~\cite{lin2022fqvit} introduces a fully-quantized method, which respectively designs Powers-of-Two Scale and Log-Int-Softmax for post-LayerNorm and post-Softmax activations.
In PTQ4ViT~\cite{yuan2022ptq4vit}, a twin uniform quantizer is introduced to handle the long-tail post-Softmax activations and uneven post-GELU activations, complemented by a Hessian-guided metric for searching quantization scales.
APQ-ViT~\cite{ding2022towards} establishes a block-wise error reconstruction and a Matthew-effect preserving quantizer for post-Softmax activations. 
In Evol-Q~\cite{frumkin2023jumping}, an evolutionary search method is employed to search extremely sensitive quantization parameters. 
RepQ-ViT~\cite{li2023repq} introduces a reparameterization technique to handle high-variant post-LayerNorm activations, where the channel-wise quantizers are simplified to layer-wise quantizers. Also, a Log$\sqrt{2}$ quantizer is adopted to accommodate post-Softmax activations.
GPTQ~\cite{frantar-gptq} employs OBS~\cite{frantar2022optimal} to progressively compensate for weight quantization error by utilizing Hessian information. 
IGQ-ViT~\cite{moon2024instance} employs instance-aware group quantization for ViTs, where activations and post-Softmax scores are split into multiple groups dynamically for each instance and each group utilizes its own quantization parameters.
OAS-ViT~\cite{maoutlier} provides theoretical insights to analyze the role of reconstruction granularity in mitigating the outlier problem in ViTs, which is also validated by the experimental results.

\section{Preliminaries}

\subsection{Quantizers}

For a fair comparison, our quantization settings are aligned with the earlier work~\cite{li2023repq}. Specifically, we quantize the weights and activations of all matrix multiplications in ViTs. The channel-wise quantizer (along the output channel) and layer-wise quantizer are adopted for weights and activations, respectively.
For weights and the activations except for the post-Softmax activations, we adopt the uniform quantizer.
Given full-precision values $\mathbf{v}$ and the bit-width $b$, the uniform quantizer is defined as:
\begin{align}
  \bar{\mathbf{v}} = \text{Q}_{un}(\mathbf{v}, b) = s \cdot \text{clip}\left(\left\lfloor \frac{\mathbf{v}}{s} \right\rceil+z, 0, 2^b-1 \right),
\label{eq:UQ}
\end{align}
where $\left\lfloor\cdot\right\rceil$ denotes the rounding function, $\text{clip}(\cdot)$ makes the output between $0$ and $2^b-1$, the scale factor $s$ is determined through a grid search aimed at minimizing the error before and after quantization, and the zero-point $z$ is calculated as $\left\lfloor-\frac{\min(\mathbf{v})}{s} \right\rceil$. Both $s$ and $z$ are quantization parameters.
For long-tail post-Softmax activations, the log$\sqrt{2}$ quantizer~\cite{li2023repq} is adopted:
\begin{align}
\label{eq:lg2}
  & \bar{\mathbf{v}}  = \text{Q}_{lg\sqrt{2}}(\mathbf{v}, b)  = s \cdot 2^{\lfloor -\frac{x_q}{2}\rfloor} (\mathbbm{1}{(\mathbf{v}_q)}(\sqrt{2} - 1)+1), \\
  & \mathbf{v}_q =  \text{clip}\left(\left\lfloor -2\text{log}_2\frac{\mathbf{v}}{s} \right\rceil, 0, 2^b-1 \right),
\end{align}
where $\mathds{1}(\cdot)$ returns 0 for even numbers and 1 for odd numbers, $s$ is determined through a grid search aimed at minimizing the error before and after quantization.
All quantization parameters for the aforementioned quantizers are determined based on the calibration datasets.

\subsection{Objective}

Denoting the full-precision activation as $\mathbf{x} \sim \mathcal{P}(\mathbf{x})$, where $\mathbf{x} \in \mathbb{R}^{D_{in}}$, and the weight $\mathbf{W} \in \mathbb{R}^{D_{out} \times D_{in}}$.  Here, $D_{in}$ and $D_{out}$ are the input and output dimensions, respectively.
The quantization error induced by activation and weight quantization is denoted as $\delta{\mathbf{x}} = \bar{\mathbf{x}}-\mathbf{x} $ and $\delta\mathbf{W} =\bar{\mathbf{W}}-\mathbf{W}$, respectively. For each layer, we aim to minimize the Mean Squared Error (MSE) before and after weight-activation quantization:
\begin{equation}
\begin{aligned}
  \mathcal{L}^{\text{MSE}} & = \mathbb{E} \left[\| \mathbf{W}\mathbf{x} - \bar{\mathbf{W}}\bar{\mathbf{x}}\|_2^2 \right] \\
  & = \mathbb{E} \left[\| \mathbf{W}\mathbf{x} - (\mathbf{W}+\delta\mathbf{W})(\mathbf{x}+\delta\mathbf{x})\|_2^2\right].
  \label{eq:obj}
\end{aligned}
\end{equation}

Eq.\,\ref{eq:obj} indicates that output error is contributed by both activations and weight quantization error.

\section{Method}

The entangled $\delta\mathbf{x}$ and $\delta\mathbf{W}$ make it a challenge to find an optimal solution for Eq.\,\ref{eq:obj}~\cite{li2021brecq}.
To make it tractable, we relax Eq.\,\ref{eq:obj} to two sequential sub-problems by respectively minimizing error from quantized activations and weights. As shown in Fig.\,\ref{fig:framework}, we first perform Activation quantization error reduction (Aqer) followed by Weight quantization error reduction (Wqer), which are respectively detailed in the following.

\begin{figure}[th]
\begin{center}
\centerline{\includegraphics[width=0.92\columnwidth]{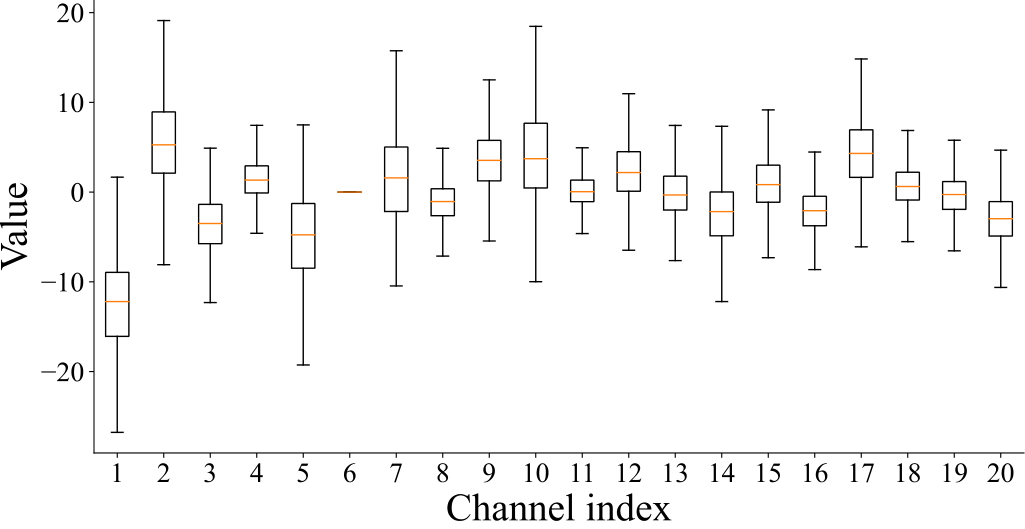}}
\caption{Example of channel distribution of activations after block.8.norm2 of DeiT-S. Results are extracted with 32 images. 
}
\label{fig:post-LayerNorm}
\end{center}
\end{figure}

\subsection{Activation Quantization Error Reduction}

We start by quantizing the activations layer-wise while retaining the weights as full-precision. Currently, given the quantization parameters, the MSE caused by activation quantization error $\delta\mathbf{x}$ is defined as the optimization target:
\begin{align}
  \mathcal{L}^{\text{MSE}} = \mathbb{E} \left[ \| \mathbf{W}\mathbf{x} - \mathbf{W}(\mathbf{x}+\delta{\mathbf{x}})\|_2^2 \right].
  \label{eq:obj-act}
\end{align}

To mitigate errors induced by activation quantization, we introduce Activation quantization error reduction (Aqer) including Reparameterization Initialization specific for high variant post-LayerNorm activations and Ridge Regression.

\subsubsection{Reparameterization Initialization}

\begin{figure}[!t]
\begin{center}
\centerline{\includegraphics[width=\columnwidth]{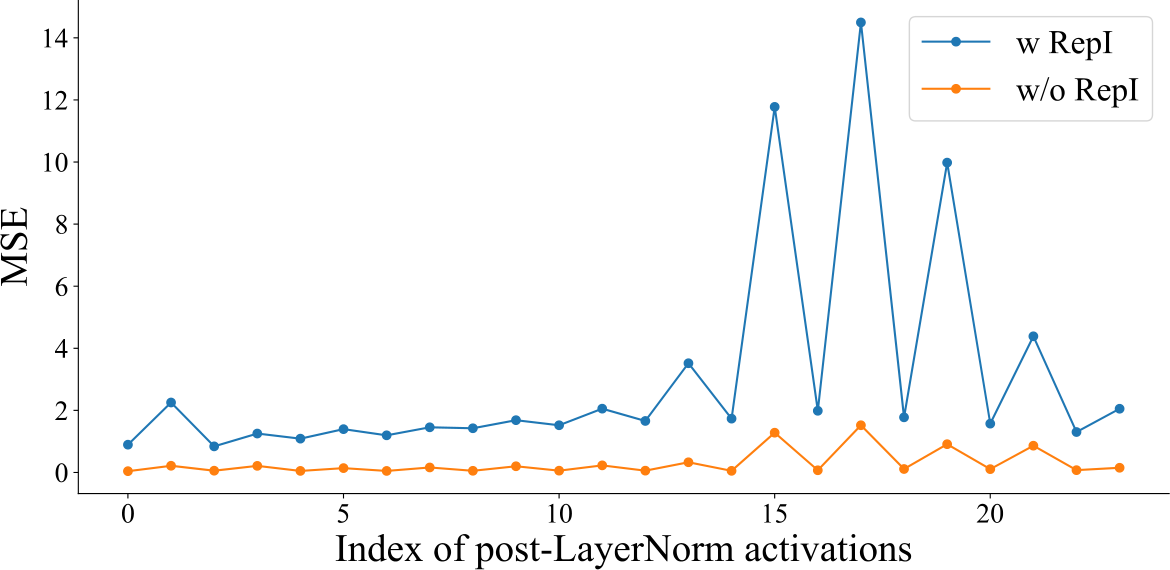}}
\caption{Comparison of MSE between using and not using Reparameterization Initialization. The MSE is evaluated by Eq.\,\ref{eq:obj-act}. ``RepI'' indicates Reparameterization Initialization. Results are derived from DeiT-S with 32 images. Activations are quantized to 4-bit. 
}
\label{fig:Rep-Initialization}
\end{center}
\end{figure}

It is evident that the MSE defined in Eq.\,\ref{eq:obj-act} decreases as $\delta\mathbf{x}$ is reduced. 
However, as illustrated in Fig.\,\ref{fig:post-LayerNorm}, post-LayerNorm activations exhibit high variance across input channels, leading to substantial quantization errors if layer-wise quantization is directly applied~\cite{lin2022fqvit}. Correspondingly, as shown in Fig.\,\ref{fig:Rep-Initialization}, a large MSE occurs.

Motivated by these observations, for post-LayerNorm activations, we adopt the reparameterization technique~\cite{li2023repq}, which first initializes a channel-wise quantizer and then translates it to the layer-wise equivalence. 
To be specific, we first initialize input channel-wise scales $\mathbf{s}\in \mathbb{R}^{D_{in}}$ and zero-point $\mathbf{z}\in \mathbb{R}^{D_{in}}$ for post-LayerNorm activations, where $D_{in}$ is the number of the input channels.
Subsequently, we adjust the parameters of LayerNorm ($\bm{\beta}\in \mathbb{R}^{D_{in}}$ and $\bm{\gamma}\in \mathbb{R}^{D_{in}}$) and the weights and biases ($\mathbf{W} \in \mathbb{R}^{D_{out} \times D_{in}} $ and $\mathbf{b} \in \mathbb{R}^{D_{in}}$) of the subsequent layer by using Eq.\,\ref{eq:raparm_1} and Eq.\,\ref{eq:raparm_2}, respectively.
\begin{equation}
    \label{eq:raparm_1}
  \bm{\beta}^{new} = \frac{\bm{\beta}+\mathbf{s}\odot \mathbf{r}_2}{\mathbf{r}_1}, \quad \bm{\gamma}^{new} = \frac{\bm{\gamma}}{\mathbf{r}_1}.
\end{equation}
\begin{equation}
\label{eq:raparm_2}
\begin{aligned}
  & \mathbf{W}_{:,j}^{new} = \mathbf{r}_1\odot\mathbf{W}_{:,j}, \mathbf{b}_j^{new} = \mathbf{b}_j - (\mathbf{s}\odot \mathbf{r}_2) \mathbf{W}_{:,j}, \\ 
  & \text{where} \quad j \in 1, 2, \dots,  D_{in}.
\end{aligned}
\end{equation}

Here, $\odot$ denotes Hadamard product, $\tilde{s}=\text{mean}(\mathbf{s}) \in \mathbb{R}^{1}$, $\tilde{z}=\text{mean}(\mathbf{z}) \in \mathbb{R}^{1}$, $\mathbf{r}_1=\mathbf{s}/\tilde{s}$, and $\mathbf{r}_2=\mathbf{z}-\tilde{z}$.
After adjustment, the channel-wise quantizer is transformed to a layer-wise quantizer.

As shown in Fig.\,\ref{fig:Rep-Initialization}, Reparameterization Initialization significantly reduces MSE for all post-LayerNorm activations within the network, thereby demonstrating its effectiveness.
As for other activations, which exhibit a relatively smooth distribution, we continue to apply layer-wise quantization~\cite {lin2022fqvit}.

\subsubsection{Ridge Regression}

After initialization, we further mitigate activation quantization error $\delta\mathbf{x}$ by formulating the Ridge Regression problem.
To be specific, we suggest minimizing Eq.\,\ref{eq:obj-act} by adding $\mathbf{W}$ with an adjustment $\delta\mathbf{W}^*$: 
\begin{equation}
\begin{aligned}
  &\mathbb{E} \left[ \| \mathbf{W}\mathbf{x} - (\mathbf{W} + \delta\mathbf{W}^*)(\mathbf{x}+\delta{\mathbf{x}})\|_2^2 \right]  + \lambda_1 \| \delta\mathbf{W}^* \|_2^2 
  \\
   & = \mathbb{E} \left[\| - \delta\mathbf{W}^*(\mathbf{x}+\delta{\mathbf{x}}) - \mathbf{W}\delta{\mathbf{x}}  \|_2^2\right] + \lambda_1 \| \delta\mathbf{W}^* \|_2^2
   \\
   & = \mathbb{E} \left[ \| \delta\mathbf{W}^*\bar{\mathbf{x}} + \mathbf{W}\delta{\mathbf{x}}  \|_2^2 \right] + \lambda_1 \| \delta\mathbf{W}^* \|_2^2.
  \label{eq:obj-act1}
\end{aligned}
\end{equation}

Here, $\delta\mathbf{W}^*$ denotes adjustment that is computed by Ridge Regression, $\bar{\mathbf{x}}=\mathbf{x}+\delta\mathbf{x}$ is the quantized input, $\lambda_1\| \delta\mathbf{W}^* \|_2^2$ acts as the regularization term, $\lambda_1$ is a hyper-parameter that controls the intensity of the regularization.
Eq.\,\ref{eq:obj-act1} constitutes the Ridge Regression problem. To minimize it, we first compute its gradient \emph{w.r.t.} $\delta\mathbf{W}^*$:
\begin{equation}
\begin{aligned}
     \frac{\partial}{\partial \delta\mathbf{W}^*} & \mathbb{E}\left[    \| \delta\mathbf{W}^*\bar{\mathbf{x}} + \mathbf{W}\delta{\mathbf{x}}  \|_2^2  \right] + \lambda_1 \| \delta\mathbf{W}^* \|_2^2 
    \\
    & =  \mathbb{E} \left[ 2 (\delta\mathbf{W}^*\bar{\mathbf{x}} + \mathbf{W}\delta{\mathbf{x}})\bar{\mathbf{x}}^T \right] + 2\lambda_1 \delta\mathbf{W}^*.
    \label{eq:obj-act2}
\end{aligned}
\end{equation}

Then, we solve for $\delta\mathbf{W}^*$ by setting Equation \ref{eq:obj-act2} to zero:
\begin{equation}
\begin{aligned}
& \mathbb{E}\left[ 2 (\delta\mathbf{W}^*\bar{\mathbf{x}} + \mathbf{W}\delta{\mathbf{x}})\bar{\mathbf{x}}^T \right] + 2\lambda_1 \delta\mathbf{W}^*  = 0
\\
& \Rightarrow  \delta\mathbf{W}^* = -\mathbf{W}  \mathbb{E} \left[\delta{\mathbf{x}}\bar{\mathbf{x}}^T\right](\mathbb{E} \left[\bar{\mathbf{x}}\bar{\mathbf{x}}^T \right] + \lambda_1 \mathbf{I})^{-1}.
\label{eq:eq10}
\end{aligned}
\end{equation}

The regularization term $\lambda_1 \mathbf{I}$ ensures the inverse of $\mathbb{E} \left[\bar{\mathbf{x}}\bar{\mathbf{x}}^T \right] + \lambda_1 \mathbf{I}$ always exists, which is crucial for computational stability. In addition, it suppresses outliers, thereby mitigating overfitting and enhancing the model's generalizability.
Suppressing outliers is also crucial for subsequent weight quantization since it restricts the range of weights. This restriction prevents the quantization points from being distributed in the uncovered region, thus enhancing the expressive ability of quantization~\cite{APoT}.
In practice, given the calibration dataset, we estimate $\mathbb{E}\left[\delta\mathbf{x}\bar{\mathbf{x}}^T\right]$ and $\mathbb{E}\left[\bar{\mathbf{x}}\bar{\mathbf{x}}^T \right]$ using $\frac{1}{N}\sum_n^N \delta{\mathbf{x}}_n\bar{\mathbf{x}}_n^T$ and $\frac{1}{N}\sum_n^N \bar{\mathbf{x}}_n\bar{\mathbf{x}}_n^T$, respectively. 
Here, $N = B\times T$, where $B$ is the size of the calibration dataset, and $T$ is the number of tokens of one image.
Note that $\delta\mathbf{x}$ and $\bar{\mathbf{x}}$ are determined given the input and the quantization parameters.
After obtaining $\delta\mathbf{W}^*$, we incorporate it into the network's weights by $\mathbf{W} = \mathbf{W} + \delta\mathbf{W}^*$.
By doing so, the proposed Aqer explicitly mitigates the quantization error from quantized activations into the weights.

\begin{figure}[ht]
\centering

\subfloat[]{%
  \includegraphics[width=0.95\linewidth]{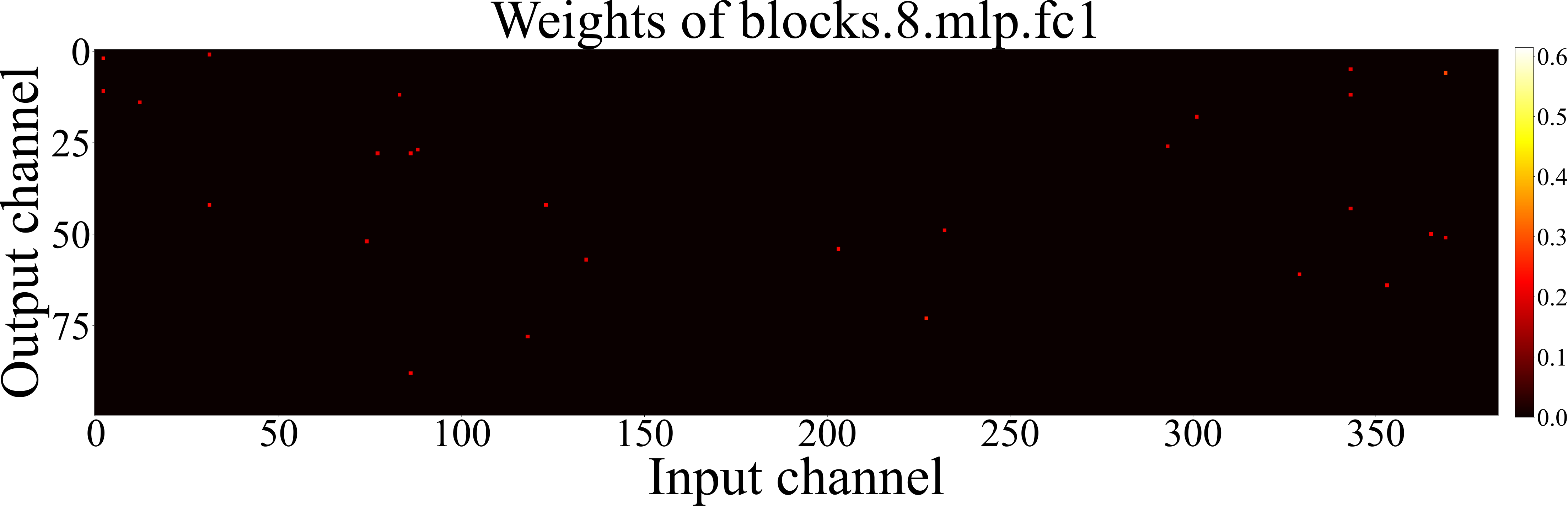}%
  \label{fig:weight-afterRI-1}
}
\hfill
\subfloat[]{%
  \includegraphics[width=0.95\linewidth]{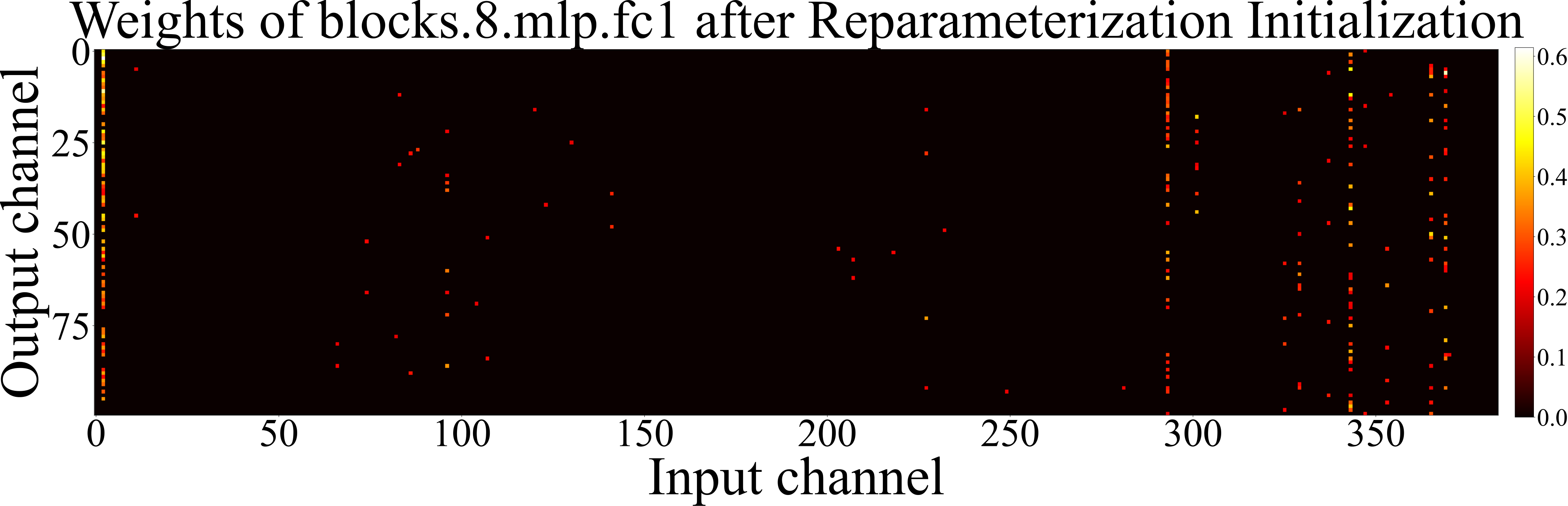}%
  \label{fig:weight-afterRI-2}
}

\caption{Heatmap of absolute weight values: (a) Before and (b) After Reparameterization Initialization. Weights are extracted from blocks.8.mlp.fc1 of DeiT-S. For better visualization, elements with absolute values less than 0.2 have been set to zero.}
\label{fig:weight-afterRI}
\end{figure}

\subsection{Weight Quantization Error Reduction}

After completing Aqer, the activation is quantized, and we proceed to quantize the weights. Here, given the quantization parameters, we solely consider the MSE caused by weight quantization error $\delta\mathbf{W}$ to obtain the
following optimization target:
\begin{equation}
\begin{aligned}
 \mathcal{L}^{\text{MSE}} & = \mathbb{E} \left[\| \mathbf{W}\bar{\mathbf{x}} - (\mathbf{W}+\delta\mathbf{W})\bar{\mathbf{x}}\|_2^2 \right] = \sum_i^{D_{out}} \mathcal{L}^{\text{MSE}}_i
 \\
 & = \sum_i^{D_{out}} \mathbb{E} \left[\| \mathbf{W}_{i,:}\bar{\mathbf{x}} - (\mathbf{W}_{i,:}+\delta\mathbf{W}_{i,:})\bar{\mathbf{x}}\|_2^2 \right].
  \label{eq:obj-weight0}
\end{aligned}
\end{equation}

Eq.\,\ref{eq:obj-weight0} suggests that the minimization across output channels is conducted independently. 
To mitigate the resulting quantization error, we introduce Weight quantization error reduction (Wqer), which includes Dual Uniform Quantization specific for weights initially set by Reparameterization Initialization, Rounding Refinement, and Ridge Regression.

\begin{figure}[th]
\begin{center}
\centerline{\includegraphics[width=\columnwidth]{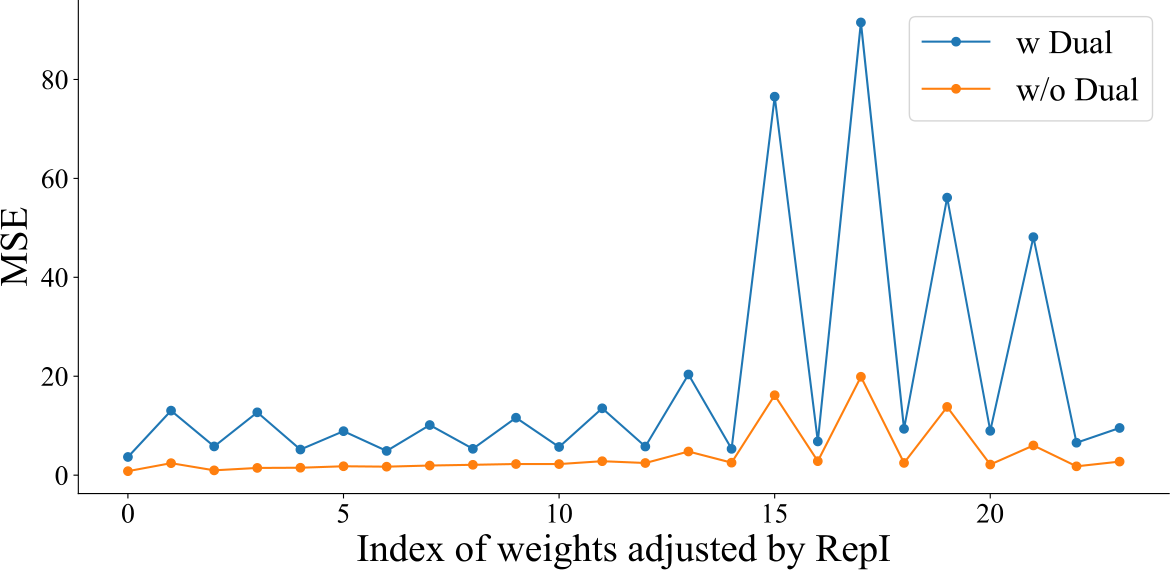}}
\caption{Comparison of MSE between using and not using Dual Uniform Quantization. The MSE is evaluated by Eq.\,\ref{eq:obj-weight0}. ``Dual'' indicates Dual Uniform Quantization. Results are derived from W4A4 DeiT-S with 32 images.
}
\label{fig:dual-quant}
\end{center}
\end{figure}

\subsubsection{Dual Uniform Quantization}

It is evident that the smaller $\delta\mathbf{W}_{i,:}$ leads a reduction in Eq.\,\ref{eq:obj-weight0}. Note that the weights are quantized channel-wise (along with the output channel), meaning that the values of $\mathbf{W}_{i,:}$ share the same quantization parameters.

However, the weights adjusted by Reparameterization Initialization tend to present a quantization-unfriendly distribution. As illustrated in Fig.\,\ref{fig:weight-afterRI-1} and Fig.\,\ref{fig:weight-afterRI-2}, after employing Reparameterization Initialization, the weights in output channels suffer from extensive outliers. The issue stems from Eq.\,\ref{eq:raparm_2}, which involves an input channel-wise multiplication of weights. Consequently, each value of $\mathbf{W}_{i,:}$ is multiplied by a factor $\mathbf{r}_1=\mathbf{s}/\tilde{s}$, which can significantly increase the values if $\mathbf{s}$ and $\tilde{s}$ diverge substantially, leading to the emergence of outliers.
Due to this uneven distribution of $\mathbf{W}_{i,:}$, using the same quantization parameters either incurs significant clip noise or rounding errors~\cite{ACIQ}, inevitably leading to substantial quantization errors $\delta\mathbf{W}_{i,:}$~\cite{fang2020post}. 
As a result, as demonstrated in Fig.\,\ref{fig:dual-quant}, a large MSE occurs.

Fortunately, as demonstrated in Fig.\,\ref{fig:weight-afterRI-2}, the occurrence of outliers typically concentrates on certain input channels.
Inspired by this observation, we introduce Dual Uniform Quantization, which employs two independent uniform quantizers to respectively manage input channels with and without outliers to reduce $\delta\mathbf{W}_{i,:}$.
To determine which channels should be selected as outlier channels, we introduce an outlier channel selection algorithm, detailed in Alg.\,\ref{alg:maximum_outlier_coverage}. This algorithm identifies a set $\mathcal{O}$, whose size $|\mathcal{O}|$ is a predefined hyper-parameter, that encompasses input channels most affected by outliers.

At first, we perform outlier detection for each output channel by utilizing percentile-based thresholds. For each output channel $i$ (where $i = 1, \dots, D_{out} $), we calculate the 99-th percentile ($ \tau_i^{99} $) and the 1-th percentile ($ \tau_i^1 $) to serve as the upper and lower thresholds, respectively. Any value that falls outside the range defined by $ \tau_i^{99} $ and $ \tau_i^{1} $ is considered an outlier:
\begin{equation}
\begin{aligned}
 \mathcal{M}_i & = \{ \mathbf{W}_{i,j} \in \mathbf{W}_{i,:} : \mathbf{W}_{i,j} < \tau_i^{1} \text{ or } \mathbf{W}_{i,j} > \tau_i^{99} \}.
\label{eq:identify-outlier}
\end{aligned}
\end{equation}

Subsequently, we perform greedy selection to form $\mathcal{O}$. In particular, for each input channel $j$ (where $ j = 1, \dots, D_{in} $), we calculate the frequency of outlier occurrences, which is given by:
\begin{equation}
\begin{aligned}
f_j = \frac{1}{D_{out}} \sum_{i=1}^{D_{out}} \mathds{1}(\mathbf{W}_{i,j} \in \mathcal{M}_i ),
\label{eq:outlier_fre}
\end{aligned}
\end{equation}
where $ \mathds{1} $ returns 1 if $w_{ij}$ is an outlier for output channel $i$, and 0 otherwise.
After calculating the frequency of outliers for all input channels, we greedy select the top $|\mathcal{O}|$ input channels with the highest frequency to form the set $\mathcal{O}$:
\begin{equation}
\begin{aligned}
\mathcal{O} = \mathrm{top\_index}(f, |\mathcal{O}|)
\label{eq:select}
\end{aligned}
\end{equation}
where $f$ is the set of $f_j$, and $\mathrm{top\_index}$ returns the index of the top $|\mathcal{O}|$ values in $f$. The time costs of Alg.\,\ref{alg:maximum_outlier_coverage} are minimal. For example, on W4A4 DeiT-S using a single NVIDIA 3090 GPU and an Intel Xeon 4214R CPU, the process takes approximately 24 seconds. In the following, we demonstrate the optimality of the above greedy selection in terms of outlier coverage.

\begin{algorithm}[tb]
    \caption{Outlier Channel Selection}
    \label{alg:maximum_outlier_coverage}
    \begin{algorithmic}[1]
        \STATE \textbf{Input:} $\mathbf{W} \in \mathbb{R}^{D_{out} \times D_{in}}$
        \STATE \textbf{Initialize:} $\mathcal{O} = \varnothing$
        \STATE \textit{/* Outlier Detection */}
        \FOR{$i = 1$ to $D_{out}$}
            \STATE Initialize $\mathcal{M}_i = \varnothing$
            \STATE Compute 99-th and 1-th percentiles $\tau_i^{99}$, $\tau_i^1$ for $\mathbf{W}_{i,:}$
            \STATE Obtain outliers set $\mathcal{M}_i$ for $\mathbf{W}_{i,:}$ by Eq.\,\ref{eq:identify-outlier}
        \ENDFOR
        \STATE \textit{/* END Outlier Detection */}
        \STATE \textit{/* Greedy Selection */}
        \STATE Calculate the frequency of outliers for each input channel $j$ by Eq.\,\ref{eq:outlier_fre}
        \STATE Select the top $|\mathcal{O}|$ input channels with the highest $f_j$ to form $\mathcal{O}$
        \STATE \textit{/* END Greedy Selection */}
        \STATE \textbf{Output:} Set $\mathcal{O}$ of selected input channels
    \end{algorithmic}
\end{algorithm}

%%%%

\begin{proposition}
Given the definition of outliers as specified in Eq.\,\ref{eq:identify-outlier}, the greedy selection algorithm achieves the maximal coverage rate of outliers in $\mathbf{W}$.
\end{proposition}

\begin{proof}
Assume for contradiction that there exists an alternative set $\mathcal{O}'$ with $|\mathcal{O}'| = |\mathcal{O}|$, which achieves better outlier coverage than the set $\mathcal{O}$ selected by the algorithm.

Define $ f_{j'} $ as the frequency of outliers for each $j'$ in $\mathcal{O}'$ and $f_j$ as the frequency for each $j$ in $\mathcal{O}$. By assumption that $\mathcal{O}'$ achieves higher coverage, it follows that:
\begin{equation}
    \sum_{j' \in \mathcal{O}'} f_{j'} > \sum_{j \in \mathcal{O}} f_j.
    \label{eq:greater_coverage}
\end{equation}

From Eq.\,\ref{eq:greater_coverage}, it follows that there exists at least one $ j' $ in $\mathcal{O}'$ not in $\mathcal{O}$ with $ f_{j'} > f_j $ for some $ j $ in $\mathcal{O}$:
\begin{equation}
\exists j' \in \mathcal{O}' \setminus \mathcal{O}, \exists j \in \mathcal{O} : f_{j'} > f_j.
\label{eq:proof-1}
\end{equation}

However, according to Eq.\,\ref{eq:select}, every $ j $ in $\mathcal{O}$ has $ f_j $ greater than or equal to any $ f_{j'} $ for $ j' $ not in $\mathcal{O}$:
\begin{equation}
\forall j \in \mathcal{O}, \forall j' \notin \mathcal{O} : f_j \geq f_{j'}.
\label{eq:condition}
\end{equation}

Eq.\,\ref{eq:proof-1} and Eq.\,\ref{eq:condition} lead to a contradiction.
Therefore, the assumption that $\mathcal{O}'$ achieves better coverage than $\mathcal{O}$ must be false. By contradiction, the set $\mathcal{O}$ selected by Eq.\,\ref{eq:select} maximizes the coverage of outliers.
\end{proof}

%%%%

The proposed outlier channel selection algorithm ensures that $\mathcal{O}$ consists of the channels most affected by outliers, thereby enabling Dual Uniform Quantization to separately quantize these specific channels that are prone to exacerbating quantization errors due to the presence of significant outliers. Specifically, after obtaining $\mathcal{O}$, for weights $\mathbf{W}_{i,:}$ of $i$-th output channel, the Dual Uniform Quantization is defined as:
\begin{align}
  & \bar{\mathbf{W}}_{i,j} = \text{Q}_{un}^1(\mathbf{W}_{i,j}, b), j \in \mathcal{O},\\
  & \bar{\mathbf{W}}_{i,j} = \text{Q}_{un}^2(\mathbf{W}_{i,j}, b), j \notin \mathcal{O}.
\end{align}

\begin{table}[!t]
\centering
\small
\caption{Results of W4A4 DeiT-S with different methods for minimizing $\mathbb{E} \left[ \| \delta\mathbf{W}^s_{i,:}\bar{\mathbf{x}}^s \|_2^2 \right]$. ``baseline'' indicates only perform Aqer. We use the pulp (a CPU-only LP modeler written in Python) to solve the MIPQ. The high time costs of using $\mathbb{E} \left[ \| \delta\mathbf{W}^s_{i,:}\bar{\mathbf{x}}^s \|_2^2 \right]$ make the results meaningless.}
% \resizebox{0.5\textwidth}{!}{%
\begin{tabular}{ccc}
\toprule[1.25pt]
\textbf{Method} & \textbf{Time costs}  &\textbf{Acc. (\%)} \\
\midrule[0.75pt]
\midrule[0.75pt] 
Baseline (Aqer) & $\sim$ 1 minute   & 70.96 \\
+ MIPQ w/  $\mathbb{E} \left[ \| \delta\mathbf{W}^s_{i,:}\bar{\mathbf{x}}^s \|_2^2 \right]$ & $\sim$130 hours  & - \\
+ MIPQ w/ Proxy & $\sim$10 hours  &  71.89 \\
+ Rounding Refinement & $\sim$4 minutes   &  71.80 \\
\bottomrule[1.0pt]
\end{tabular}
% }
\label{tab:timecompar}
\end{table}

\begin{figure}[ht]
\begin{center}
\centerline{\includegraphics[width=1\columnwidth]{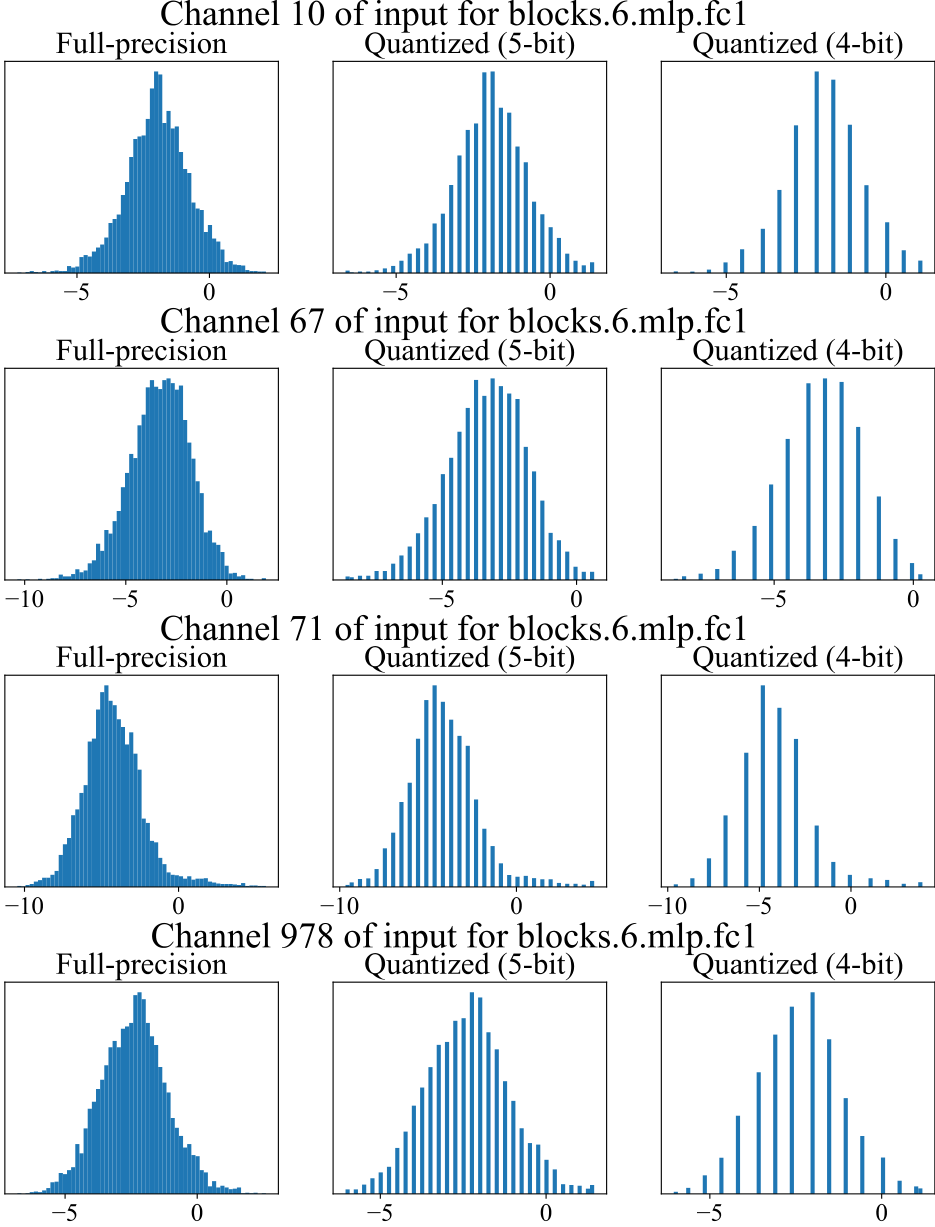}}
\caption{Distribution of activations across different channels. Results are extracted from W4A4 DeiT-S with 32 images.}
\label{fig:distribution_plot}
\end{center}
\end{figure}

Here, the extra overhead of Dual Uniform Quantization is negligible. For example, this only leads to less than 1\% extra storage costs for W4A4 DeiT-S. 
Moreover, since $\mathcal{O}$ is pre-defined before inference, an efficient forward propagation can be achieved by reordering the input channels of the weights and activations as discussed in \cite{yuan2023rptq}.
As shown in Fig.\,\ref{fig:dual-quant}, implementing Dual Uniform Quantization significantly reduces the Mean Squared Error (MSE).
For the weights of other layers that are less affected by outliers, we continue to apply standard uniform quantization.

\subsubsection{Rounding Refinement}

Afterward, we further mitigate the weight quantization error $\delta\mathbf{W}_{i,:}$.
Simultaneously quantizing the entire full-precision weights yields unrecoverable quantization error~\cite{frantar-gptq}. 
Thus, we adopt an iterative quantization-and-correction manner to gradually minimize quantization error~\cite{Zhou2017Incremental}. In each iteration, the first half of unquantized weights is quantized, followed by a mitigation of the resulting quantization error.
Specifically, we begin with the current full-precision weight $\mathbf{W}_{i,:}$ and the corresponding $\bar{\mathbf{x}}$. 
We then partition $\mathbf{W}$ into two segments: the first half, $\mathbf{W}^s_{i,:} \in \mathbb{R}^{ 1\times D_{in}^s}$, is designated for quantization, while the remaining part, $\mathbf{W}^r_{i,:} \in \mathbb{R}^{1 \times D_{in}^r}$, is retained at full-precision.  
Correspondingly, we derive $\bar{\mathbf{x}}^s \in \mathbb{R}^{D_{in}^s}$ and $\bar{\mathbf{x}}^r \in \mathbb{R}^{D_{in}^r}$ from $\bar{\mathbf{x}}$, where $\bar{\mathbf{x}}^s$ and $\bar{\mathbf{x}}^r$ respectively contain the rows of $\bar{\mathbf{x}}$ corresponding to $\mathbf{W}^s_{i,:}$ and $\mathbf{W}^r_{i,:}$.
The quantization error of the quantized $\mathbf{W}^s_{i,:}$ is denoted as $\delta\mathbf{W}^s_{i,:} = \bar{\mathbf{W}}^s_{i,:} - \mathbf{W}^s_{i,:}$, and the resulting MSE is:
\begin{equation}
\begin{split}
 \mathcal{L}^{\text{MSE}}_i & = \mathbb{E} \big[ \| [ \mathbf{W}^s_{i,:},\mathbf{W}^r_{i,:} ] [ \bar{\mathbf{x}}^s, \bar{\mathbf{x}}^r ] 
 \\
 & \quad\quad\quad -  [ \mathbf{W}^s_{i,:}+\delta\mathbf{W}^s_{i,:},\mathbf{W}^r_{i,:} ] [ \bar{\mathbf{x}}^s, \bar{\mathbf{x}}^r ] \|_2^2 \big] 
 \\
 & = \mathbb{E} \left[ \| \delta\mathbf{W}^s_{i,:}\bar{\mathbf{x}}^s \|_2^2 \right].
\end{split}
\label{eq:obj-weight-divide}
\end{equation}

Here, $\mathbf{W}_{i,:} = [ \mathbf{W}^s_{i,:},\mathbf{W}^r_{i,:} ]$, $\bar{\mathbf{x}} = [ \bar{\mathbf{x}}^s, \bar{\mathbf{x}}^r ]$. To mitigate Eq.\,\ref{eq:obj-weight-divide}, we first introduce Rounding Refinement, in which the rounding direction of the quantized weights is refined, i.e., adjusting $\delta\mathbf{W}^s_{i,:}$, to reduce $\mathbb{E} \left[ \| \delta\mathbf{W}^s_{i,:}\bar{\mathbf{x}}^s \|_2^2 \right]$ itself. 
Then, given $\mathbb{E} \left[ \| \delta\mathbf{W}^s_{i,:}\bar{\mathbf{x}}^s \|_2^2 \right]$ after Rounding Refinement, we formulate a Ridge Regression problem to further mitigate it by adjusting $\mathbf{W}^r_{i, :}$.

\begin{figure}[ht]
\begin{center}
\centerline{\includegraphics[width=\columnwidth]{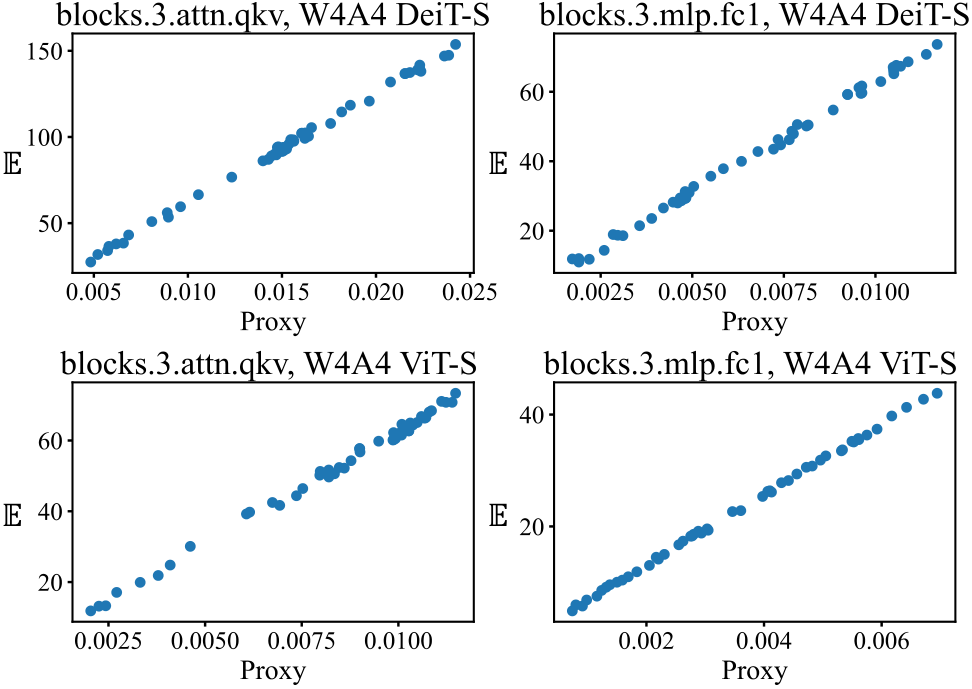}}
\caption{$\mathbb{E}$ denotes $\mathbb{E} \left[ \| \delta\mathbf{W}^s_{i,:}\bar{\mathbf{x}}^s \|_2^2 \right]$. The proxy values are positively correlated with the real values. 
}
\label{fig:proxyvsreal}
\end{center}
\end{figure}

At first, we aim to adjust the rounding direction of quantized weights to minimize $\mathbb{E} \left[ \| \delta\mathbf{W}^s_{i,:}\bar{\mathbf{x}}^s \|_2^2 \right]$.
Specifically, for the $j$-th value in $\mathbf{W}^s_{i,:}$, denoted as $\mathbf{W}^s_{i,j}$, the quantization involves rounding it either to the floor or ceil~\cite{Upordown}.
Thereby the quantization error for $\mathbf{W}^s_{i,:}$, denoted as $\delta\mathbf{W}^s_{i,j}$, can be represented as either $\delta\mathbf{W}^{s\downarrow}{i, j}$ or $\delta\mathbf{W}^{s\uparrow}{i, j}$.
Here, $\delta\mathbf{W}^{s\downarrow}_{i, j} = \mathbf{W}^s_{i,j} - \text{Q}_{un\downarrow}(\mathbf{W}^s_{i,j}, b) > 0$ denotes error from rounding-to-floor strategy, $\delta\mathbf{W}^{s\uparrow}_{i, j} = \mathbf{W}^s_{i,j} - \text{Q}_{un\uparrow}(\mathbf{W}^s_{i,j}, b) < 0 $ denotes error from rounding-to-ceil strategy, where $\downarrow/\uparrow$ denote replacing $\left\lfloor \cdot \right\rceil$ in Eq.\,\ref{eq:UQ} with $\left\lfloor \cdot \right\rfloor$/$\left\lceil \cdot \right\rceil$. The selection of $ \delta\mathbf{W}^s_{i,:}$ is an NP-hard problem, whose solution can be searched by the mixed-integer quadratic program (MIPQ)~\cite{pia2017mixed,kuzmin2023pruning}. 
However, the high computational complexity of $\mathbb{E} \left[ \| \delta\mathbf{W}^s_{i,:}\bar{\mathbf{x}}^s \|_2^2 \right]$ makes it a challenge to find the solution with reasonable time costs. As shown in Tab.\,\ref{tab:timecompar}, using $\mathbb{E} \left[ \| \delta\mathbf{W}^s_{i,:}\bar{\mathbf{x}}^s \|_2^2 \right]$ as the target of MIPQ consumes prohibitive $\sim$130 hours, which makes the results meaningless.

\begin{algorithm}[ht]
       \caption{Weight Quantization Error Reduction}
        \label{alg:wqer}
        \begin{algorithmic}[1]
           \STATE {\bfseries Input:} $\mathbf{W} \in \mathbb{R}^{D_{out} \times D_{in}}$, $\{\bar{\mathbf{x}}_n \in \mathbb{R}^D_{in}\}_{n=1}^N$, maximum iteration $\text{T}$
           \STATE \textbf{Initialize:} $\bar{\mathbf{W}} = \varnothing$
           \FOR{$i = 1$ to $D_{out}$}
               \STATE $\bar{\mathbf{W}}_{i, :} = \varnothing$, $\{\bar{\mathbf{x}}_n\}_{n=1}^N = \{\bar{\mathbf{x}}_n\}_{n=1}^N$
               % \STATE \textbf{while} $|\mathbf{W}_{i, :}| > 0$
               \WHILE{$|\mathbf{W}_{i, :}| > 0$}
                   \STATE  Partition $\mathbf{W}_{i, :}$ into $\big[\mathbf{W}^s_{i, :},\mathbf{W}^r_{i, :}\big]$
                   \STATE Partition $\{\bar{\mathbf{x}}_n\}_{i=1}^N$ into \{$\big[\bar{\mathbf{x}}_n^s,\bar{\mathbf{x}}_n^r\big]\}_{i=1}^N$ 
                   \STATE  \textit{/* Rounding Refinement */}
                   \STATE  Obtain $\hat{\boldsymbol{\mu}^s}\hat{\boldsymbol{\mu}}^{sT} + \hat{\boldsymbol{\Sigma}}^s$ from cache or calculate it with $\{\bar{\mathbf{x}}_n^s\}_{n=1}^N$, calculate $\delta\mathbf{W}^{s\downarrow}_{i, :}$, $\delta\mathbf{W}^{s\uparrow}_{i, :}$ with $\mathbf{W}^s_{i, :}$
                       % \STATE \quad \textbf{while} $0 \leq \text{T-}\text{-}$: 
                       \WHILE{$0 \leq \text{T-}\text{-}$}
                           \STATE Calculate proxy $\mathcal{L}_{old}$ with $\delta\mathbf{W}^s_{i, :}$ by Eq.\,\ref{eq:obj-weight3}
                           \STATE Calculate gradients $\boldsymbol{G}_{\delta\mathbf{W}^s_{i, :}}$ by Eq.\,\ref{eq:obj-weight4}
                           \STATE  Obtain $\mathcal{S}$ by Eq.\,\ref{eq:obj-weight5}
                           \STATE  Obtain adjusted $\delta\mathbf{W}^{'}_{i, :}$ by Eq.\,\ref{eq:obj-weight6}
                           \STATE  Calculate proxy $\mathcal{L}_{now}$ with $\delta\mathbf{W}^{'}_{i, :}$ by Eq.\,\ref{eq:obj-weight3}
                           \IF{$\mathcal{L}_{now} > \mathcal{L}_{old}$}
                           \STATE break
                           \ENDIF
                           \STATE  $\delta\mathbf{W}^s_{i, :} = \delta\mathbf{W}^{'}_{i, :}$
                    \ENDWHILE
                    \STATE  $\bar{\mathbf{W}}_{i, :} \leftarrow \bar{\mathbf{W}}_{i, :} \cup (\mathbf{W}^s_{i, :} + \delta\mathbf{W}^s_{i, :})$
                    \STATE \textit{/* END Rounding Refinement */}
                \ENDWHILE
                \STATE \textit{/* Ridge Regression */}
                \STATE Calculate $\delta\mathbf{W}^{r*}_{i, :}$ by Eq.\,\ref{eq:obj-steptwosolution}
                \STATE  $\mathbf{W}_{i, :} \leftarrow \mathbf{W}^r_{i, :} + \delta\mathbf{W}^{r*}_{i, :}$
                \STATE \textit{/* END Ridge Regression */}
                \STATE  $\{\bar{\mathbf{x}}_n\}_{n=1}^N \leftarrow \{\bar{\mathbf{x}}_n^r\}_{n=1}^N$
               \STATE $\bar{\mathbf{W}} \leftarrow \bar{\mathbf{W}} \cup \bar{\mathbf{W}}_{i, :}$
        \ENDFOR
        \STATE {\bfseries Output:} Quantized weight $\bar{\mathbf{W}}$
        \end{algorithmic}
\end{algorithm}

\textbf{Efficient Proxy}.
Therefore, we aim to find an efficient proxy for $\mathbb{E} \left[ \| \delta\mathbf{W}^s_{i,:}\bar{\mathbf{x}}^s \|_2^2 \right]$. First, we re-write $\mathbb{E} \left[ \| \delta\mathbf{W}^s_{i,:}\bar{\mathbf{x}}^s \|_2^2 \right]$ as:
\begin{equation}
\begin{aligned}
 \mathbb{E} \left[ \| \delta\mathbf{W}^s_{i,:}\bar{\mathbf{x}}^s \|_2^2 \right] &  \overset{\Delta}{=} (\mathbb{E} \left[ \delta\mathbf{W}^s_{i,:}\bar{\mathbf{x}}^s  \right])^2 + \text{Var} \left[ \delta\mathbf{W}^s_{i,:}\bar{\mathbf{x}}^s  \right].
  \label{eq:obj-weight1}
\end{aligned}
\end{equation}

Here, $\Delta$ indicates utilizing $\mathbb{E}\left[ Z^2 \right] = (\mathbb{E}\left[ Z \right])^2 + \text{Var}\left[ Z \right]$.
As proved by \cite{klambauer2017self}, according to the central limit theorem, the numerous multiplication and addition operations within neural networks make the activations generally follow a Gaussian distribution, which is also a basic assumption in many previous works in the quantization field~\cite{ding2019regularizing,sun2022entropy,lin2022fqvit,chmiel2020robust}.
Meanwhile, Fig.\,\ref{fig:distribution_plot} illustrates the channel distribution of the full-precision and quantized activations. Here, the 67th and 71th channels are influenced by outliers, resulting in a larger range compared to the other channels. However, it can be observed that both outlier and non-outlier channels exhibit quantized activations that approximate a Gaussian distribution~\cite{whitepaper}. Note that the quantization process inherently clips outliers, thereby their impact on the distribution is mitigated largely.

Based on the above analysis, we consider channel distribution of $\bar{\mathbf{x}}^s$ still can be captured by the Gaussian distribution, and model $\bar{\mathbf{x}}^s$ with a $D_{in}^s$-dimensional Gaussian distribution $\mathcal{N}(\boldsymbol{\mu}^s, \boldsymbol{\Sigma}^s)$,
where $D_{in}^s$ is the dimension of $\bar{\mathbf{x}}^s$, $\boldsymbol{\mu}^s \in \mathbb{R}^{D_{in}^s}, \boldsymbol{\Sigma}^s \in \mathbb{R}^{D_{in}^s \times D_{in}^s}$. 
Then, the Eq.\,\ref{eq:obj-weight1} becomes:
\begin{equation}
\begin{aligned}
     & \mathbb{E} \left[ \delta\mathbf{W}^s_{i,:}\bar{\mathbf{x}}^s  \right]^2  + \text{Var} \left[ \delta\mathbf{W}^s_{i,:}\bar{\mathbf{x}}^s  \right]
     \\ 
     & \quad =   \delta\mathbf{W}^s_{i,:}\boldsymbol{\mu}^s\boldsymbol{\mu}^{sT}(\delta\mathbf{W}^s_{i,:})^T + \delta\mathbf{W}_{i,:}\boldsymbol{\Sigma}^s(\delta\mathbf{W}^s_{i,:})^T
    \\
    & \quad = \delta\mathbf{W}^s_{i,:}(\boldsymbol{\mu}^s\boldsymbol{\mu}^{sT} + \boldsymbol{\Sigma}^s)(\delta\mathbf{W}^s_{i,:})^T.
    \label{eq:obj-weight3}
\end{aligned}
\end{equation}

\begin{table*}[!ht]
\centering
\small
\caption{Results on ImageNet dataset. The Top-1 accuracy (\%) is reported as the metric. ``W/A'' indicates that the bit-width of the weights and activations are W and A bits, respectively. ``*'' indicates the results are re-produced by using the official code.}
% \resizebox{0.9\textwidth}{!}{%
\begin{tabular}{ccccccccc}
\toprule[1.25pt]
\textbf{Method}  & \textbf{W/A} & \textbf{ViT-S}& \textbf{ViT-B} & \textbf{DeiT-T}& \textbf{DeiT-S} & \textbf{DeiT-B} &  \textbf{Swin-S}& \textbf{Swin-B} \\
\midrule[0.75pt]
\midrule[0.75pt]
Full-Precision   & 32/32 & 81.39 & 84.54 & 72.21 & 79.85 & 81.80 & 83.23 & 85.27 \\ 
\midrule[0.75pt]
FQ-ViT*~\cite{lin2022fqvit}  & 3/4 & 0.10 & 0.10 & 0.10 & 0.10 & 0.10 & 0.10  & 0.10 \\
PTQ4ViT*~\cite{yuan2022ptq4vit} & 3/4 & 0.10 & 0.10 & 0.20 & 0.15 &0.59 & 0.64 & 0.53  \\
GPTQ*~\cite{frantar-gptq} &  3/4 & 23.32 & 44.63 & 42.25 & 48.95 & 61.75 & 66.71 & 71.43 \\
DuQuant*~\cite{lin2024duquant} &  3/4 & 0.40 & 1.24 &9.25 & 8.53 & 47.29 & 70.70 & 54.68   \\
SpinQuant*~\cite{liu2024spinquant}  & 3/4 &  22.38 &  0.50 & 29.94 &  49.56  &  18.06 &  35.83 &  14.67  \\
 OmniQuant*~\cite{shaoomniquant} &  3/4 &   0.32 &  0.35 &  2.45 &  9.67   &  23.91 &  65.98 &  52.05  \\
 SmoothQuant*~\cite{xiao2023smoothquant}  &  3/4  &   0.23 &  0.27 &  4.60 &  10.72    &  25.28  &  68.51 &  53.70 \\
RepQ-ViT*~\cite{li2023repq} &  3/4 & 15.65 & 26.98 & 29.34 & 45.82 & 58.92 & 59.83 & 44.17  \\
AdaRound*~\cite{nagel2020up} &  3/4 &	11.04 &		4.72	 &	36.05	 &	33.56 &	62.50 &		68.12	 &	53.92 \\
BRECQ*~\cite{li2021brecq} &   3/4 & 4.97 & 1.25 & 29.23 & 18.58 & 40.49 & 66.93 & 53.38 \\
QDrop*~\cite{wei2021qdrop} &  3/4  & 9.77  & 11.87  & 17.85  & 30.27  & 61.12  & 73.47  & 74.33\\
PD-Quant*~\cite{liu2023pd} &  3/4 & 4.56 & 21.81 & 41.87 & 41.65 & 53.63 & 70.07 & 56.48  \\
OAS-ViT*~\cite{maoutlier} &  3/4 & 3.56 & 7.43 & 32.34 & 36.68 & 51.51 & 72.37 & 73.50 \\
ERQ (Ours) &  3/4 &  \underline{\textbf{60.13}} &   \underline{\textbf{72.37}} &  \underline{\textbf{52.73}} &   \underline{\textbf{69.09}} &  \underline{\textbf{76.47}} &   \underline{\textbf{78.12}} &  \underline{\textbf{79.98}} \\
\midrule[0.75pt]
FQ-ViT~\cite{lin2022fqvit} &   4/4 & 0.10 & 0.10 & 0.10 & 0.10 & 0.10 & 0.10  & 0.10 \\
PTQ4ViT~\cite{yuan2022ptq4vit} &  4/4 & 42.57 & 30.69 & 36.96 & 34.08 & 64.39  & 76.09 & 74.02 \\
APQ-ViT~\cite{ding2022towards} & 4/4 & 47.95 & 41.41 & 47.94 & 43.55 & 67.48 & 77.15 & 76.48 \\
GPTQ*~\cite{frantar-gptq} &  4/4 & 67.59 & 75.12 & 58.96 & 70.85 & 76.10 & 80.17 & 81.08  \\
DuQuant*~\cite{lin2024duquant} &  4/4 &  1.73 &  9.53 &22.68 & 26.57 & 64.87 & 78.59 & 78.56\\
SpinQuant*~\cite{liu2024spinquant}  &  4/4 &  50.41 &  5.42 & 42.57 &  58.44  &  60.67 &  64.16  &  50.80 \\
 OmniQuant*~\cite{shaoomniquant}  &  4/4 &    8.80 & 5.50 &  10.67 &  16.85   &  38.74 &  77.48 &  78.14  \\
 SmoothQuant*~\cite{xiao2023smoothquant}  &  4/4 &   13.19 & 8.22 &  21.78 &  32.76   &  49.31 &  79.33 &  79.26   \\
RepQ-ViT~\cite{li2023repq} & 4/4 & 65.05 & 68.48 &57.43 & 69.03 & 75.61  & 79.45 & 78.32 \\
AdaRound*~\cite{nagel2020up} &   4/4 &	63.09	& 70.51	& 55.65	& 69.24	& 75.20	& 76.05 &	78.12 \\
BRECQ*~\cite{li2021brecq} &  4/4 & 11.31 & 3.03 & 38.41  & 32.89  & 59.10  & 68.40   & 56.51 \\
QDrop*~\cite{wei2021qdrop} &  4/4 & 17.77 & 21.72 & 31.65 & 35.79 & 65.47 & 78.92  & 80.49\\
PD-Quant*~\cite{liu2023pd} &  4/4 & 32.64 & 34.86 & 58.50 & 64.85 & 60.06 & 77.04 & 75.84  \\
ERQ (Ours) &  4/4 &  \underline{\textbf{71.61}} &  \underline{\textbf{78.65}} &  \underline{\textbf{61.79}} &  \underline{\textbf{74.35}} &  \underline{\textbf{79.18}} &  \underline{\textbf{81.19}} &  \underline{\textbf{83.32}} \\
\bottomrule[1.0pt]
\end{tabular}
% }
\label{tab:imagenet-0}
\end{table*}

Here, Eq.\,\ref{eq:obj-weight3} is the obtained proxy for $\mathbb{E} \left[ \| \delta\mathbf{W}^s_{i,:}\bar{\mathbf{x}}^s \|_2^2 \right]$. 
In practice, we estimate the empirical $\hat{\boldsymbol{\mu}}^s$ and $\hat{\boldsymbol{\Sigma}}^s$ with the given calibration dataset. Note that for all output channels, $\hat{\boldsymbol{\mu}}^s$ and $\hat{\boldsymbol{\Sigma}}^s$ are shared and require only a single computation.
Fig.\,\ref{fig:proxyvsreal} presents the relationship between the proxy and $\mathbb{E} \left[ \| \delta\mathbf{W}^s_{i,:}\bar{\mathbf{x}}^s \|_2^2 \right]$. It can be seen that the proposed proxy is proportional to the real value, demonstrating its fidelity.

The computational complexity of using our proxy is $O((D_{in}^s)^2)$, while the complexity of $\mathbb{E} \left[ \| \delta\mathbf{W}^s_{i,:}\bar{\mathbf{x}}^s \|_2^2 \right]$ is $O(ND_{in}^s)$, where $N >> D_{in}^s$.
Thus, the proxy can serve as a low-cost objective for solving $\delta\mathbf{W}^s_{i,:}$.
As shown in Tab.\,\ref{tab:timecompar}, using Eq.\,\ref{eq:obj-weight3} as the target of MIPQ reduces the time costs from $\sim$130 hours to $\sim$10 hours. 
However, this still incurs moderate costs since current open-source implementations of MIPQ only support CPU and cannot fully exploit the capacity of GPU.
In the next, we introduce Rounding Refinement, a GPU-supported method that uses the gradient of the proxy to adjust $\delta\mathbf{W}^s_{i,:}$ faster.

\begin{table*}[!ht]
\centering
\small
\caption{Results on ImageNet dataset. ``-'' denotes the original paper does not report corresponding results and the code is unavailable.}
% \resizebox{0.9\textwidth}{!}{%
\begin{tabular}{ccccccccc}
\toprule[1.25pt]
\textbf{Method}  & \textbf{W/A} & \textbf{ViT-S}& \textbf{ViT-B} & \textbf{DeiT-T}& \textbf{DeiT-S} & \textbf{DeiT-B} &  \textbf{Swin-S}& \textbf{Swin-B} \\
\midrule[0.75pt]
\midrule[0.75pt]
Full-Precision   & 32/32 & 81.39 & 84.54 & 72.21 & 79.85 & 81.80 & 83.23 & 85.27 \\ 
\midrule[0.75pt]
FQ-ViT*~\cite{lin2022fqvit} & 5/5 &0.10 &0.10 &0.10 &0.10 &0.10 &0.10 &0.10  \\
PTQ4ViT*~\cite{yuan2022ptq4vit} &5/5 & 72.74  & 72.32 & 65.00 & 70.26 & 72.65 & 80.90 & 81.87 \\
GPTQ*~\cite{frantar-gptq} &   5/5 & 78.63 & 82.06 & 69.05 & 77.12 & 80.17 &  82.19 &  83.00  \\
DuQuant*~\cite{lin2024duquant} &   5/5 &  67.87 &  75.33 & 60.89 &  74.11 &  78.72 &  82.06 &  84.08 \\
SpinQuant*~\cite{liu2024spinquant} &   5/5 &  71.30  &  72.19 & 60.76 &  70.69 &  77.27 &   82.17 &  82.50  \\
 OmniQuant*~\cite{shaoomniquant}  &   5/5 &  56.33 &  61.99 & 42.57 &  64.13 &  76.43 &  81.44 &  83.66  \\
 SmoothQuant*~\cite{xiao2023smoothquant} &   5/5 & 67.70&  71.88 & 56.63 &  71.24 &  77.83 &  81.89 &  84.01  \\
RepQ-ViT*~\cite{li2023repq} &  5/5 & 78.43 &82.03 &69.00 &77.04 &80.08 &82.08 &83.22 \\
AdaRound*~\cite{nagel2020up} & 5/5 &	77.53 &	82.00 &	68.87 &	76.22 &	80.18	 &82.12	 &84.09\\
BRECQ*~\cite{li2021brecq} & 5/5 & 47.35 & 43.51 & 62.12 & 63.15 & 75.61 & 80.66 & 82.31\\
QDrop*~\cite{wei2021qdrop} & 5/5 &  56.32 & 57.92 & 62.36 & 70.07 & 78.41 & 81.73 & 83.61 \\
PD-Quant*~\cite{liu2023pd}  & 5/5 &  65.06 & 58.40 & 68.02 & 74.94 & 74.61 & 81.27 & 82.12 \\
OAS-ViT*~\cite{maoutlier} &  5/5 & 40.40 & 47.13 & 62.34 &  65.96 & 77.36 & 80.91 & 83.60 \\
ERQ (Ours) &   5/5 &  \underline{\textbf{79.13}} &  \underline{\textbf{82.99}} &  \underline{\textbf{69.76}} &  \underline{\textbf{77.98}} &  \underline{\textbf{80.92}} &  \underline{\textbf{82.49}} &  \underline{\textbf{84.72}} \\
\midrule[0.75pt]
FQ-ViT~\cite{lin2022fqvit}  & 6/6 & 4.26 & 0.10 & 58.66 & 45.51 & 64.63  & 66.50 & 52.09 \\
PSAQ-ViT~\cite{li2022patch}  & 6/6 & 37.19 & 41.52 & 57.58 & 63.61 & 67.95   & 72.86 & 76.44 \\
Ranking~\cite{liu2021post}  & 6/6 & - & 75.26 & - & 74.58 & 77.02 &   - & - \\
PTQ4ViT~\cite{yuan2022ptq4vit}  & 6/6 & 78.63 & 81.65 & 69.68 & 76.28 & 80.25  & 82.38 & 84.01 \\
APQ-ViT \cite{ding2022towards} & 6/6 & 79.10 & 82.21 & 70.49 & 77.76 & 80.42  & 82.67 & 84.18 \\
NoisyQuant$\dagger$~\cite{liu2023noisyquant}  & 6/6 & 76.86  & 81.90 & - & 76.37  & 79.77 &    82.78 & 84.57 \\
NoisyQuant$\ddagger$~\cite{liu2023noisyquant}  & 6/6 & 78.65  & 82.32 & - & 77.43 & 80.70 &   82.86 & 84.68 \\
GPTQ*~\cite{frantar-gptq} & 6/6 & 80.44 & 83.72 & 71.05 & 78.95 & 81.37 & 82.82 & 84.89  \\
DuQuant*~\cite{lin2024duquant} &   6/6 &  79.13 &  82.73 & 68.95 &  78.46 &  81.06 &  82.74 &  84.88   \\
SpinQuant*~\cite{liu2024spinquant}  &   6/6 &  78.96 &  82.74 & 69.69 &  76.80 &  80.65 &  82.00 &    84.47   \\
 OmniQuant*~\cite{shaoomniquant}  &   6/6 &  75.23 &  78.06 & 62.39 &  76.62 &  80.10 &  82.21 &  84.63  \\
 SmoothQuant*~\cite{xiao2023smoothquant} &   6/6 &  78.09 &  82.33 & 69.31 &  78.67 &  80.87 &  82.99 &  85.00  \\
RepQ-ViT~\cite{li2023repq} &  6/6 & 80.43 & 83.62 & 70.76 & 78.90 & 81.27 &   82.79 &84.57 \\
EasyQuant~\cite{wu2020EasyQuant} &   6/6 & 75.13 & 81.42 & - & 75.27 & 79.47 &   82.45 & 84.30 \\
Bit-shrinking~\cite{lin2023bit} & 6/6 & 80.44 & 83.16 & - & 78.51 & 80.47 & 82.44 & - \\
BRECQ*~\cite{li2021brecq} &    6/6 & 61.18 & 71.29 & 69.62 & 70.93 & 79.46 & 81.85 & 84.08\\
QDrop*~\cite{wei2021qdrop} &  6/6 & 68.57 & 74.38 & 69.98 & 76.57 & 80.66 & 82.53 & 84.31\\
PD-Quant*~\cite{liu2023pd} &   6/6  & 71.38 & 63.14 & 70.74 & 77.63 & 79.32 & 82.33 & 84.38 \\
OAS-ViT*~\cite{maoutlier} &  6/6 & 61.32 & 71.87 & 69.67 & 71.52 & 79.87 & 82.16 & 84.26 \\
ERQ (Ours) & 6/6 &  \underline{\textbf{80.68}} &  \underline{\textbf{84.00}} &  \underline{\textbf{71.24}}&  \underline{\textbf{79.14}} &   \underline{\textbf{81.54}}&  \underline{\textbf{82.94}} &  \underline{\textbf{85.06}} \\
\bottomrule[1.0pt]
\end{tabular}
% }
\label{tab:imagenet-1}
\end{table*}

\textbf{Rounding Refinement}.
At first, we initialize $\delta\mathbf{W}^s_{i,j}$ with the rounding-to-nearest strategy. Now, $\delta\mathbf{W}^s_{i,j}$ is either equal to $\delta\mathbf{W}^{s\downarrow}_{i, j}$ or $\delta\mathbf{W}^{s\uparrow}_{i, j}$. Then, we aim to determine an index set $\mathcal{S}$ that contains the index set of the elements necessitating modifications, whose rounding direction is overturned:
\begin{equation}
\begin{aligned}
    \delta\mathbf{W}_{i, j}^s = 
    \begin{cases} 
        \delta\mathbf{W}^{s\downarrow}_{i, j} & \text{if} \,\, \delta\mathbf{W}_{i, j}^s = \delta\mathbf{W}^{s\uparrow}_{i, j} \\
        \delta\mathbf{W}^{s\uparrow}_{i, j} & \text{otherwise.}
    \end{cases}
    , j \in \mathcal{S}.
    \label{eq:obj-weight6}
\end{aligned}
\end{equation}

To determine $\mathcal{S}$, we first take the derivative of the proxy (Eq.\,\ref{eq:obj-weight3}) \emph{w.r.t} the $\delta\mathbf{W}^s_{i,:}$
\begin{equation}
\begin{aligned}
     \boldsymbol{G}_{\delta\mathbf{W}^s_{i,:}} & =  \frac{\partial}{\partial \delta\mathbf{W}^s_{i,:}} \delta\mathbf{W}^s_{i,:}(\boldsymbol{\mu}^s\boldsymbol{\mu}^{sT} + \boldsymbol{\Sigma}^s)(\delta\mathbf{W}^s_{i,:})^T \\
     & = 2 \delta\mathbf{W}^s_{i,:}(\boldsymbol{\mu}^s\boldsymbol{\mu}^{sT} + \boldsymbol{\Sigma}^s). 
    \label{eq:obj-weight4}
\end{aligned}
\end{equation}

We only select the elements whose gradients are the same sign, since this is the only way to allow overturn. For example, given $\delta\mathbf{W}_{i, j}^s = \delta\mathbf{W}^{^s\downarrow}_{i, j}$, replacing it by $\delta\mathbf{W}^{^s\uparrow}_{i, j}$ is feasible only if $\boldsymbol{G}_{\delta\mathbf{W}_{i, j}^s}$ has the same sign as $\delta\mathbf{W}_{i, j}^s$. 
Thus, the index set $\mathcal{S}$ is defined as:
\begin{equation}
\begin{aligned}
     & \mathcal{S} =   \mathrm{top\_index}(\mathcal{M}, \mathrm{k}), 
     \\
     & \mathcal{M} =  \lvert \boldsymbol{G}_{\delta\mathbf{W}_{i, :}^s} \odot \mathds{1}(\boldsymbol{G}_{\delta\mathbf{W}_{i, :}^s} \odot \delta\mathbf{W}_{i, :}^s ) \rvert \in \mathbb{R}^{D_{in}^s}.
    \label{eq:obj-weight5}
\end{aligned}
\end{equation}

Here, $\mathrm{top\_index}$ returns the index of the top $\mathrm{k}$ elements, $\mathds{1}(\cdot)$ returns 1 for non-negative input and 0 for negative input, $\lvert \cdot \rvert$ returns the absolute value of the input. 
After obtaining $\mathcal{S}$, the overturn is performed with Eq.\,\ref{eq:obj-weight6}. The above process iterates until the adjusted $\delta\mathbf{W}^s_{i, :}$ incurs a larger proxy value or reaches maximum iterations \text{T}. After obtaining $\delta\mathbf{W}^s_{i, :}$, the quantization can be completed by $\bar{\mathbf{W}}^s_{i, :} = \mathbf{W}^s_{i, :}+\delta\mathbf{W}^s_{i, :}$. $\bar{\mathbf{W}}^s_{i, :}$ is then added into the set of quantized weights.
The overall process of Rounding Refinement is presented in Lines 7 - Lines 18 of Alg.\,\ref{alg:wqer}.
As shown in Tab.\,\ref{tab:timecompar}, Rounding Refinement significantly reduces the time costs from 10 hours to 4 minutes by 150$\times$ at the cost of affordable accuracy loss. 

\subsubsection{Ridge Regression}

After Rounding Refinement, we suggest adjusting $\mathbf{W}^r_{i, :}$ with $\delta\mathbf{W}^{r*}_{i, :}$ to further counteract $\mathbb{E} \left[ \| \delta\mathbf{W}^s_{i,:}\bar{\mathbf{x}}^s \|_2^2 \right]$, which yields the following target:
\begin{equation}
\begin{split}
 % \mathcal{L}^{\text{MSE}}_i 
 % = \mathbb{E} \big[ \| [ \mathbf{W}^s_{i, :},\mathbf{W}^r_{i, :} ] [ \bar{\mathbf{x}}^s, \bar{\mathbf{x}}^r ] 
 % \\
 % & \quad\quad -  [ \mathbf{W}^s_{i, :}+\delta\mathbf{W}^s_{i, :},\mathbf{W}^r_{i, :}+\delta\mathbf{W}^{r*}_{i, :}] [ \bar{\mathbf{x}}^s, \bar{\mathbf{x}}^r] \|_2^2 \big]
 % % \\
  \mathbb{E} \big[  \|\delta\mathbf{W}^s_{i, :}\bar{\mathbf{x}}^s +  \delta\mathbf{W}^{r*}_{i, :}\bar{\mathbf{x}}^r \|_2^2 \big]  + \lambda_2\| \delta\mathbf{W}^{r*}_{i, :} \|_2^2,
\end{split}
\label{eq:obj-weight7}
\end{equation}
where $\lambda_2$ is a hyper-parameter to control intensity of the regularization term $\lambda_2\| \delta\mathbf{W}^{r*}_{i, :} \|_2^2$. The minimization of Eq.\,\ref{eq:obj-weight7} formulates the Ridge Regression problem and the solution is defined as:
\begin{equation}
\begin{split}
 \delta\mathbf{W}^{r*}_{i, :} = - \delta\mathbf{W}^s_{i, :}\mathbb{E} \left[ \bar{\mathbf{x}}^s\bar{\mathbf{x}}^{rT} \right](\mathbb{E} \left[ \bar{\mathbf{x}}^r \bar{\mathbf{x}}^{rT} \right] + \lambda_2 \mathbf{I})^{-1}.
\end{split}
\label{eq:obj-steptwosolution}
\end{equation}

In practice, we estimate $\mathbb{E}\left[\bar{\mathbf{x}}^r \bar{\mathbf{x}}^{sT}\right]$ and $\mathbb{E}\left[\bar{\mathbf{x}}^r \bar{\mathbf{x}}^{rT} \right]$ by using $\frac{1}{N}\sum_n^N \bar{\mathbf{x}}_n^r\bar{\mathbf{x}}_n^{sT}$ and $\frac{1}{N}\sum_n^N \bar{\mathbf{x}}_n^r\bar{\mathbf{x}}_n^{rT}$. 
Afterward, $\mathbf{W}^r_{i, :} = \mathbf{W}^r_{i, :}+\delta\mathbf{W}^{r*}_{i, :}$ to mitigate the error. Currently, $\mathbf{W}^r_{i, :}$ remains as full-precision and will be processed in the next iteration. Such a process continuously runs until all weights are accurately quantized. The overall process of proposed Rounding Refinement and Ridge Regression is presented in Alg.\,\ref{alg:wqer}. In practice, we perform Rounding Refinement and Ridge Regression for multiple output channels in parallel, thereby achieving a highly efficient implementation.

\begin{table*}[!ht]
\centering
\small
\caption{Results on COCO dataset. ``AP$^\text{box}$'' denotes the box average precision for object detection, and ``AP$^\text{mask}$'' denotes the mask average precision for instance segmentation. ``*'' and ``$\dagger$''  indicate the results are re-produced by using the official code.}
% \resizebox{0.9\textwidth}{!}{%
\begin{tabular}{cccccccccc}
\toprule[1.25pt]
\multirow{3.5}{*}{\textbf{Method}}  & \multirow{3.5}{*}{\textbf{W/A}} & \multicolumn{4}{c}{\textbf{Mask R-CNN}} & \multicolumn{4}{c}{\textbf{Cascade Mask R-CNN}} \\
\cmidrule(lr){3-6}\cmidrule(lr){7-10}
&& \multicolumn{2}{c}{\textbf{w. Swin-T}} & \multicolumn{2}{c}{\textbf{w. Swin-S}} & \multicolumn{2}{c}{\textbf{w. Swin-T}} & \multicolumn{2}{c}{\textbf{w. Swin-S}} \\
&& AP$^\text{box}$ & AP$^\text{mask}$ & AP$^\text{box}$ & AP$^\text{mask}$ & AP$^\text{box}$ & AP$^\text{mask}$ & AP$^\text{box}$ & AP$^\text{mask}$ \\
\midrule[0.75pt]
\midrule[0.75pt]
Full-Precision &  32/32 & 46.0 & 41.6 & 48.5 & 43.3 & 50.4 & 43.7 & 51.9 & 45.0 \\
\midrule[0.75pt]
GPTQ*~\cite{frantar-gptq} &  3/4 & 22.9 & 25.0 & 31.7 & 32.5 & 39.8 & 35.8 & 44.6 & 39.6 \\
DuQuant*~\cite{lin2024duquant} &  3/4 &   2.0 & 2.4 &  23.5 &  26.4 &  3.7 & 3.9 &  35.7 &  35.3 \\  
SpinQuant*~\cite{liu2024spinquant} &  3/4 &  1.1 &  1.4 &  16.9 &  19.8 &  4.3 & 4.4 &  33.2 &  33.6   \\
 OmniQuant*~\cite{shaoomniquant} &  3/4 &   10.9 &  12.4 &  8.6 &  9.8 &  19.8 & 18.2 &  14.8 &  13.6   \\
 SmoothQuant*~\cite{xiao2023smoothquant} &  3/4 &   10.8 &  12.2 &  8.5 &  9.6 &  19.7 & 18.0 &  14.7 &  13.5 \\
RepQ-ViT*~\cite{li2023repq} &  3/4 & 22.2 & 24.1 & 27.9 & 29.9 &  40.2 & 35.7 & 43.7 & 38.8 \\
AdaRound*~\cite{Upordown} &   3/4 & 2.8 & 4.2 & 5.5 & 7.5 & 20.3 & 22.6 & 21.3 & 24.0 \\
BRECQ*~\cite{li2021brecq} &   3/4 & 14.7 & 18.2 & 15.3 & 19.5 & 29.9 & 28.0 & 32.3 & 29.3  \\
QDrop*~\cite{wei2021qdrop} &  3/4 & 8.7 & 10.5 &  27.3 & 30.8  & 21.6 & 19.7 &  41.3 & 37.0\\
PD-Quant*~\cite{liu2023pd} &   3/4 & 3.7 & 4.6 & 3.8 & 4.5 & 1.0 & 9.2 & 6.4 & 6.1 \\
OAS-ViT*~\cite{maoutlier} & 3/4 & 8.5 & 8.2 & 9.6 & 11.9 & 16.7 & 12.0 & 21.6 &  19.6 \\
ERQ (Ours) & 3/4 &  \underline{\textbf{27.2}} &  \underline{\textbf{28.9}} &  \underline{\textbf{30.6}} &  \underline{\textbf{33.0}} &  \underline{\textbf{45.1}} &  \underline{\textbf{40.0}} &  \underline{\textbf{47.3}} &  \underline{\textbf{42.0}}  \\
\midrule[0.75pt]
PTQ4ViT~\cite{yuan2022ptq4vit} &   4/4 & 6.9 & 7.0 & 26.7 & 26.6 & 14.7 & 13.5 & 0.5 & 0.5  \\
APQ-ViT~\cite{ding2022towards} & 4/4 & 23.7 & 22.6 & \textbf{44.7} & 40.1 & 27.2 & 24.4 & 47.7 & 41.1 \\
GPTQ*~\cite{frantar-gptq} &  4/4 & 36.3 & 36.3  & 42.9 & 40.2 & 47.1 & 41.5 & 49.2 & 43.2 \\
DuQuant*~\cite{lin2024duquant} &  4/4 &   5.1 & 5.1 &  41.2 &  38.9 &  8.2 & 8.0 &  41.7 &  38.1 \\
SpinQuant*~\cite{liu2024spinquant} &  4/4 &  5.1 &  5.4 &  36.6 &  34.5 &  9.3 & 14.7 &  39.5 &  39.7 \\
 OmniQuant*~\cite{shaoomniquant}  &  4/4 &  25.7 &  25.2 &  20.6 &  19.9 &  31.5 & 28.1 &  28.0 &  25.3  \\
 SmoothQuant*~\cite{xiao2023smoothquant} &  4/4 &  25.6 &  25.2 &  20.6 &  19.9 &  31.8 & 28.2 &  28.2 &  25.5  \\
RepQ-ViT~\cite{li2023repq} &  4/4 & 36.1 & 36.0 & 44.2$_{42.7}\dagger$ & 40.2$_{40.1}\dagger$ & 47.0 & 41.4 & 49.3 & 43.1 \\
AdaRound*~\cite{Upordown} &   4/4 & 16.3	& 19.8 & 22.3 &	22.5 & 34.6	& 33.4 & 35.8 &	34.5\\
BRECQ*~\cite{li2021brecq} &   4/4 & 25.2 & 27.3 & 32.4 & 32.9 & 40.4  &  35.9 & 41.5 & 37.2 \\
QDrop*~\cite{wei2021qdrop} &  4/4 & 10.4 & 11.3 & 39.7 & 37.8 & 17.9  & 16.2 & 20.1 & 17.4  \\
PD-Quant*~\cite{liu2023pd} &   4/4 & 15.7 & 16.1 & 30.2  & 28.4 & 34.5  & 30.1 &  38.6 & 34.1 \\
OAS-ViT*~\cite{maoutlier} & 4/4 & 21.7 & 23.3 & 25.1 & 26.0 & 31.5 & 28.5 & 32.6 & 29.3 \\
ERQ (Ours) & 4/4 &   \underline{\textbf{39.4}} &  \underline{\textbf{37.5}} & 44.0 &  \underline{\textbf{40.8}} &  \underline{\textbf{48.0}} &  \underline{\textbf{42.1}} &  \underline{\textbf{50.3}} &  \underline{\textbf{43.8}} \\
\midrule[0.75pt]
GPTQ*~\cite{frantar-gptq} &  5/5 & 43.3 & 40.1 & 46.3 & 42.0 & 49.4 & 43.0 & 50.8 & 44.0 \\
DuQuant*~\cite{lin2024duquant} &  5/5 &   26.9 & 25.4 &  46.8 &  42.4 & 43.4 & 37.8 &  51.1 &  44.3 \\
SpinQuant*~\cite{liu2024spinquant} &  5/5 &   30.6 & 29.0 & 46.6 & 42.2 & 42.0 & 36.6 & 50.8 &  44.2 \\
 OmniQuant*~\cite{shaoomniquant} &  5/5 &   43.1 & 39.4 &  44.2 &  39.8 &  48.2 & 42.1 &  48.6 &  42.0   \\
 SmoothQuant*~\cite{xiao2023smoothquant} &  5/5 &   43.1 & 39.5 &  44.1 &  39.8 &  48.5 & 42.2 &  48.5 &  42.0   \\
RepQ-ViT*~\cite{li2023repq} &  5/5 & 43.3 & 40.2 & 46.3 & 42.4 & 49.6 & 43.2 & 51.1 & 44.3 \\
AdaRound*~\cite{Upordown} &   5/5 & 34.5 & 35.6 & 41.0  & 39.8 & 46.8 & 42.1 & 48.4 & 43.3 \\
BRECQ*~\cite{li2021brecq} &   5/5 & 41.9 & 38.8 & 44.9 & 40.9 & 	47.8 & 41.7 & 49.6 & 43.0\\
QDrop*~\cite{wei2021qdrop} &  5/5 & 43.9 & 40.2 & 46.7 & 42.5 & 49.0 & 42.7 & 49.1 & 42.7  \\
PD-Quant*~\cite{liu2023pd} &   5/5 & 12.7 & 12.1 & 20.4 & 19.1 & 19.7 & 17.4 & 22.2 & 19.5 \\
OAS-ViT*~\cite{maoutlier} & 5/5 & 40.6 & 37.5 & 40.7 & 37.4 & 46.5 & 40.6 & 46.3 & 40.3 \\
ERQ (Ours) & 5/5 &  \underline{\textbf{44.2}} &  \underline{\textbf{40.8}} &  \underline{\textbf{47.1}} &  \underline{\textbf{42.8}} &  \underline{\textbf{49.8}} &  \underline{\textbf{43.3}} &  \underline{\textbf{51.4}} &  \underline{\textbf{44.6}}  \\
\bottomrule[1.0pt]
\end{tabular}
% }
\label{tab:coco}
\end{table*}

\section{Experiments}

\subsection{Implementation details}

\subsubsection{Models and tasks}

We conduct extensive experiments across various tasks, including image classification, object detection, instance segmentation, and image super-resolution.

For the image classification task, we evaluate our ERQ on the ImageNet dataset~\cite{russakovsky2015imagenet} using different ViT variants including ViT-S, ViT-B~\cite{DosovitskiyZ21An}, DeiT-T, DeiT-S, DeiT-B~\cite{touvron2021training}, as well as Swin-S and Swin-B~\cite{liu2021swin}. The pre-training full-precision models are downloaded from Timm Library~\cite{rw2019timm}.

For object detection and instance segmentation tasks, ERQ is evaluated using Mask R-CNN~\cite{he2017mask} and Cascade Mask R-CNN~\cite{cai2018cascade}, both of which adopt Swin-T or Swin-S~\cite{liu2021swin} as their backbones. The pre-training full-precision models are downloaded from the official repository of Swin Transformer~\cite{liu2021swin}.
These models are tested on the COCO dataset~\cite{lin2014microsoft}. We report the box average precision for object detection and the mask average precision for instance segmentation.

For the image super-resolution task, we evaluate ERQ on SwinIR~\cite{liang2021swinir}. The upscaling factors are $\times$2 and $\times$4, respectively. The pre-training full-precision models are downloaded from official repository of SwinIR~\cite{liang2021swinir}.
The models are tested on four standard benchmarks including Set5~\cite{bevilacqua2012low}, Set14~\cite{ledig2017photo}, BSD100~\cite{martin2001database} and Urban100~\cite{huang2015single}. We report the PSNR and SSIM~\cite{wang2004image} over the Y channel as the metrics.

\subsubsection{Hyperparameters}

Consistent with previous work~\cite{li2023repq}, we quantize all weights and activations involved in matrix multiplication. For the image classification task, we randomly select 32 images from the ImageNet. For object detection and instance segmentation tasks, we randomly select 1 image from the COCO dataset. As for the image super-resolution task, we randomly sample 8 images from the training set of DIV2K~\cite{timofte2017ntire}, using a training patch size of 48, identical to the size used in pretraining.

In our experiments, the quantization parameters are determined by forwarding the calibration datasets, after which these parameters are fixed. 
The values for $\mathrm{k}$ and the maximum iteration \text{T} of Rounding Refinement are set to 1 and 20, respectively, while $|\mathcal{O}|$ is empirically chosen to be 5\% of $D_{out}$.

For the image classification task, we set $\lambda_1=\lambda_2=1e4$ for ViT, DeiT-S, DeiT-B, and Swin-S/B, and $\lambda_1=\lambda_2=1e3$ for DeiT-T.
For detection and segmentation tasks, we set $\lambda_1=\lambda_2=1e4$ for Mask R-CNN and Cascade Mask R-CNN with Swin-T as the backbone, and $\lambda_1=\lambda_2=1e5$ for Mask R-CNN with Swin-S as the backbone.
For the image super-resolution task, we set $\lambda_1=\lambda_2=1e2$ for SwinIR$\times$2 and $\lambda_1=\lambda_2=1e3$ for SwinIR$\times$4.

All experiments are implemented using the PyTorch framework~\cite{paszke2019pytorch}, with a single NVIDIA 3090 GPU and an Intel Xeon 4214R CPU.

\subsubsection{Compared methods}

For these methods that originally did not report the results on ViTs such as BRECQ~\cite{li2021brecq}, QDrop~\cite{wei2021qdrop}, PD-Quant~\cite{liu2023pd},  DuQuant~\cite{lin2024duquant}, SpinQuant~\cite{liu2024spinquant}, OmniQuant~\cite{shaoomniquant}, SmoothQuant~\cite{xiao2023smoothquant}, and GPTQ~\cite{frantar-gptq}, we re-implement their results by using the official code with the same quantization settings and calibration dataset as the same ours. 
It is worth emphasizing that computational invariance is crucial for the rotation operations used in DuQuant and SpinQuant. However, due to the LN and the GELU function damaging the computational invariance, the rotation operations of DuQuant and SpinQuant cannot directly accommodate ViTs. Thus, we re-distributed their rotation operations within each ViT block. Note that both DuQuant and SpinQuan introduce extract computation costs since they utilize on-the-fly matrix multiplication before activation quantization.
Moreover, the initial implementation of GPTQ did not involve activation quantization. Thus, we employed the same quantizer as our own to quantize the activations for them, including the reparameterization technique and the log$\sqrt{2}$ quantizer. For OAS-ViT~\cite{maoutlier} that employs the token-wise activations quantization, we re-implement it with the same quantization granularity as ours for a fair comparison. 
We do not compare our ERQ with AdaLog~\cite{wu2024adalog} and IGQ-ViT~\cite{moon2024instance} as the former employs a log-based quantizer, which differs from the uniform quantizer used in this paper, and the latter adopts instance-wise group quantization, which has a different quantization granularity from our method.
For other PTQ of ViT methods including FQ-ViT~\cite{lin2022fqvit}, PSAQ-ViT~\cite{li2022patch}, Ranking~\cite{liu2021post},  PTQ4ViT~\cite{yuan2022ptq4vit}, APQ-ViT \cite{ding2022towards}, NoisyQuant~\cite{liu2023noisyquant}, RepQ-ViT~\cite{li2023repq}, EasyQuant~\cite{wu2020EasyQuant}, and Bit-shrinking~\cite{lin2023bit}, we use the result reported in their paper if it exists, otherwise, we re-implement based on their official code, also with the same calibration datasets as the same ours.

\begin{table*}[ht]
\centering
\small
\caption{PSNR/SSIM results of the compared methods and our ERQ in quantizing SwinIR~\cite{liang2021swinir} of scale $\times$2 and $\times$4.}
% \resizebox{0.9\textwidth}{!}{%
\begin{tabular}{cccccccccc}
\toprule[1.25pt]
\textbf{Model} & \textbf{Method} & \textbf{W/A} & Set5~\cite{bevilacqua2012low} & Set14~\cite{ledig2017photo} & BSD100~\cite{martin2001database} & Urban100~\cite{huang2015single} \\ \midrule[0.75pt] \midrule[0.75pt]
\multirow{14}{*}{\begin{tabular}[c]{@{}c@{}}SwinIR\\ $\times$2\end{tabular}} 
& Full-Precision &  32/32 & 38.35/0.9620 & 34.14/0.9227 & 32.44/0.9030 & 33.40/0.9393 \\ \cline{2-7}\cline{2-7}
\cline{2-7}
& BRECQ*~\cite{li2021brecq} & 4/4 & 30.91/0.8182 & 28.97/0.7756 & 27.50/0.7940 & 28.46/0.7549  \\
& QDrop*~\cite{wei2021qdrop} & 4/4 & 31.56/0.8781 & 29.30/0.8397 & 27.41/0.8528    & 28.94/0.8199  \\
& AdaRound*~\cite{Upordown} & 4/4 & 31.33/0.8281 & 29.29/0.7883 & 27.68/0.8022 & 28.63/0.7650 \\
& PTQ4SR*~\cite{tu2023toward} & 4/4 & 29.97/0.7746 & 28.08/0.7412   & 26.88/0.7583  & 27.79/0.7233  \\
& RepQ-ViT*~\cite{li2023repq} & 4/4 & 29.48/0.7727 &  27.94/0.7389   & 26.48/0.7600 &  27.42/0.7186 \\     
& GPTQ*~\cite{frantar-gptq} & 4/4 & 30.70/0.8928 & 29.27/0.8593 & 27.46/0.8706 & 29.12/0.8424 \\
& DuQuant*~\cite{lin2024duquant} &  4/4 &    31.97/0.8997 &  29.09/0.8221 &  25.35/0.7170 &   28.12/0.7688 \\
&  SpinQuant*~\cite{liu2024spinquant} &  4/4 &  25.21/0.8468  &  25.90/0.8189  &   23.59/0.8144  &   26.45/0.8024  \\
&  OmniQuant*~\cite{shaoomniquant}  &   4/4 &    25.83/0.8600 &    25.51/0.8173 &     22.94/0.8221 &     26.06/0.8032 \\
&  SmoothQuant*~\cite{xiao2023smoothquant}  &  4/4 &   25.82/0.8596 &  25.49/0.8166 &  22.96/0.8222  &  26.10/0.8052 \\
&  PD-Quant*~\cite{liu2023pd} &  4/4 & 27.79/0.8469 & 27.64/0.8223 & 25.93/0.8293 & 27.83/0.8083 \\ 
&  OAS-ViT* &  4/4 & 31.52/0.8384   & 29.45/0.7970   &  27.68/0.8092 &  28.75/0.7739 \\
& ERQ (Ours) & 4/4 &  \underline{\textbf{32.12/0.9010}} &  \underline{\textbf{29.97/0.8615}} &  \underline{\textbf{28.23/0.8726}} &  \underline{\textbf{29.66/0.8479}} \\ 
\midrule[0.75pt]
\multirow{14}{*}{\begin{tabular}[c]{@{}c@{}}SwinIR\\ $\times$4\end{tabular}} 
& Full-Precision &  32/32 & 32.72/0.9021 & 28.94/0.7914 & 27.83/0.7459 & 27.07/0.8164 \\ \cline{2-7}\cline{2-7}
\cline{2-7}
& BRECQ*~\cite{li2021brecq} & 4/4 & 27.44/0.7036 & 25.47/0.6234  & 23.40/0.6106 & 25.29/0.5955 \\
& QDrop*~\cite{wei2021qdrop} & 4/4 & 28.34/0.7630   & 26.13/0.6690   & 23.63/0.6226    & 25.80/0.6354 \\
& AdaRound*~\cite{Upordown} & 4/4 & 26.87/0.6794 & 25.14/0.6056 & 23.15/0.5925 & 25.06/0.5813 \\
& PTQ4SR*~\cite{tu2023toward} & 4/4 & 23.37/0.5190 & 21.99/0.4926  &  20.53/0.4826 & 21.50/0.4884 \\
& RepQ-ViT*~\cite{li2023repq} & 4/4 & 27.59/0.7027 & 25.67/0.6263 & 23.61/0.6185 &  25.35/0.5939  \\
& GPTQ*~\cite{frantar-gptq} & 4/4 & 26.66/0.7897 & 25.76/0.7024 & 23.76/0.6991 & 25.79/0.6687  \\
& DuQuant*~\cite{lin2024duquant} &  4/4 &   26.27/0.7296 &  24.37/0.6045 &  21.24/0.5149 &   24.35/0.5677 \\
& SpinQuant*~\cite{liu2024spinquant} &  4/4 &   22.10/0.7263 &   23.00/0.6532 &   21.04/0.6343  &   23.50/0.6199 \\
&  OmniQuant*~\cite{shaoomniquant} &  4/4 &  21.57/0.7383 &   22.56/0.6669 
 &   20.43/0.6496   &   23.18/0.6379 \\
&  SmoothQuant*~\cite{xiao2023smoothquant} &  4/4 &   21.56/0.7381 &   22.56/0.6662 &   20.43/0.6492 &   23.17/0.6375 \\
&  PD-Quant*~\cite{liu2023pd} &  4/4 & 23.78/0.7326 & 24.38/0.6644 & 22.48/0.6502 &  24.63/0.6306 \\ 
&  OAS-ViT*~\cite{maoutlier} &  4/4 & 27.91/0.7356 & 25.73/0.6418 & 23.58/0.6369 & 25.44/0.6140 \\
& ERQ (Ours) & 4/4 &  \underline{\textbf{28.64/0.8107}} &  \underline{\textbf{26.34/0.7118}} &  \underline{\textbf{24.03/0.7040}} &  \underline{\textbf{26.07/0.6782}}\\ \bottomrule[1.0pt]
\end{tabular}
% }
\label{tab:sr}
\end{table*}

\subsection{Results on ImageNet Dataset}

The results on the ImageNet dataset are presented in Tab.\,\ref{tab:imagenet-0} and Tab.\,\ref{tab:imagenet-1}.
It can be seen that the proposed ERQ showcases advantages over the compared methods in all bit-width settings, especially in the low-bit cases.
Specifically, due to the small amount of the calibration dataset, many methods typically suffer from the overfitting problem and exhibit unstable performance. 
For instance, for the W3A4 case, QDrop and PD-Quant obtain 9.77\% and 4.56\% on ViT-S, respectively.
In contrast, the proposed ERQ shows stable improvements across all variants.
Notably, ERQ demonstrates 36.81\% and 27.74\% performance gains on ViT-S and ViT-B, 10.48\%, 19.53\% and 13.97\% gains on DeiT-T, DeiT-S, and DeiT-B, 4.65\% and 5.65\% gains on Swin-S and Swin-B.
When it comes to the W4A4 case, ERQ respectively obtains 4.02\% and 3.53\% performance gains on ViT-S and ViT-B, 2.83\%, 3.50\% and 3.08\% performance gains on DeiT-T, DeiT-S, and DeiT-B, 1.02\% and 2.24\% performance gains on Swin-S and Swin-B.
In the W5A5 case, ERQ also presents the best performance. In particular, ERQ improves the performance by 0.50\% and 0.93\% on ViT-S and ViT-B, 0.71\%, 0.86\%, and 0.74\% performance gains on DeiT-T, DeiT-S, and DeiT-B, 0.30\% and 0.63\% performance gains on Swin-S and Swin-B.
For the W6A6 case, ERQ still outperforms other methods and provides a close performance with the full-precision model. For example, ERQ respectively presents only 0.26\% and 0.21\% accuracy loss compared with the full-precision model for Deit-B and Swin-B.

\subsection{Results on COCO Dataset}

The results of object detection and instance segmentation are reported in Tab.\,\ref{tab:coco}.
Specifically, in the W3A4 case, ERQ showcases notable performance improvements by improving the box AP and mask AP by 5.0 and 4.8 for Mask R-CNN with Swin-T, 2.7 and 2.2 for Mask R-CNN with Swin-S, 4.9 and 4.2 for Cascade Mask R-CNN with Swin-T, and 0.7 and 2.4 for Cascade Mask R-CNN with Swin-S.
For the W4A4 case, ERQ augments the box AP and mask AP by 3.1 and 1.2 for Mask R-CNN with Swin-T, 0.9 and 0.6 for Cascade Mask R-CNN with Swin-T, and 1.0 and 0.6 for Cascade Mask R-CNN with Swin-S.
For the W5A5 case, ERQ also provides the best performance. For instance, ERQ increases the box AP and mask AP by 0.9 and 0.6 for Mask R-CNN with Swin-T and 0.3 and 0.3 for Mask R-CNN with Swin-S.

\begin{table*}[ht]
\centering
\hspace{-15mm}
% \resizebox{\textwidth}{!}{
\begin{minipage}{.3\textwidth}
  \centering
  \small
  \caption{Ablation studies of Aqer and Wqer. ``Aqer'' and ``Wqer'' represent Activation quantization error reduction and Weight quantization error reduction, respectively. ``baseline'' indicates only performing calibration.}
  \begin{tabular}{cc|c}
  \toprule[1.25pt]
  \textbf{Aqer} & \textbf{Wqer} & \textbf{Top-1 Acc. (\%)} \\
  \midrule[0.75pt]
  \midrule[0.75pt]
   \multicolumn{2}{c|}{Baseline}     &  40.38 \\\midrule[0.75pt]
  $\checkmark$ &  & 70.96 (+30.58)  \\ %72.946
  & $\checkmark$ &  50.92 (+10.54)  \\
  $\checkmark$ & $\checkmark$ &  \underline{\textbf{74.35}} (+27.97) \\\bottomrule[1.0pt]
  \end{tabular}
  \label{tab:abl-components}
\end{minipage}
\begin{minipage}{.3\textwidth}
  \centering
  \small
  \caption{Ablations on components of Aqer. ``baseline'' indicates only perform Wqer. ``RepI'' and ``Ridge'' indicate Reparameterization Initialization and Ridge Regression, respectively.}
  \begin{tabular}{cc|c}
  \toprule[1.25pt]
  \multicolumn{2}{c|}{\textbf{Aqer}} & \multirow{2}{*}{\textbf{Top-1 Acc. (\%)}} \\ \cline{1-2}
     RepI       & Ridge   &                     \\
  \midrule[0.75pt]
  \midrule[0.75pt]
   \multicolumn{2}{c|}{Baseline (Wqer)}     &  50.92 \\\midrule[0.75pt]
  $\checkmark$ &  & 72.95 (+22.03) \\
   & $\checkmark$ &  59.59 (+8.67) \\
   $\checkmark$ & $\checkmark$ &   \underline{\textbf{74.35}} (+23.43) \\\bottomrule[1.0pt]
  \end{tabular}
  \label{tab:abl-Aqer}
\end{minipage}
\begin{minipage}{.3\textwidth}
  \centering
  \small
  \caption{Ablations on components of Wqer. ``baseline'' indicates only perform Aqer. ``Dual'', ``Rounding'', and ``Ridge'' represent Dual Uniform Quantization, Rounding Refinement, and Ridge Regression, respectively.}
  \begin{tabular}{ccc|c}
  \toprule[1.25pt]
   \multicolumn{3}{c|}{\textbf{Wqer}} & \multirow{2}{*}{\textbf{Top-1 Acc. (\%)}} \\ \cline{1-3}
     Dual  & Rounding         & Ridge         &                    \\
  \midrule[0.75pt]
  \midrule[0.75pt]
   \multicolumn{3}{c|}{Baseline (Aqer)}     &  70.96 \\\midrule[0.75pt]
  $\checkmark$ &  & & 73.03 (+2.07) \\
   & $\checkmark$ & &  71.80 (+0.84) \\
  &  & $\checkmark$ & 72.12 (+1.16) \\
  $\checkmark$ & $\checkmark$ & & 73.70 (+2.74)\\
  $\checkmark$ &  & $\checkmark$ & 74.06 (+3.10) \\
   & $\checkmark$  & $\checkmark$ & 72.76 (+1.80) \\
  $\checkmark$ & $\checkmark$  & $\checkmark$ &   \underline{\textbf{74.35}} (+3.39)  \\\bottomrule[1.0pt]
  \end{tabular}
  \label{tab:abl-Wqer}
\end{minipage}
% }
\end{table*}

\begin{figure*}[ht]
\begin{center}
\centerline{\includegraphics[width=\linewidth]{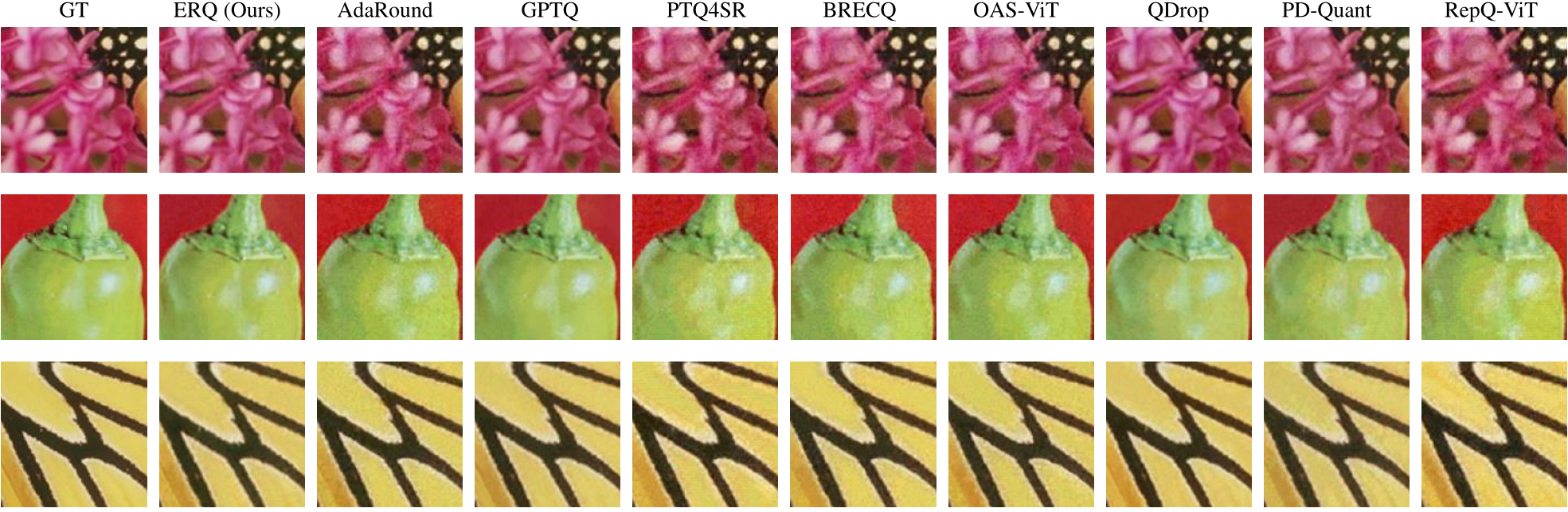}}
\caption{Qualitative comparison on the image super-resolution task.}
\label{fig:sr-results}
\end{center}
\end{figure*}

\subsection{Results on DIV2K}

The results of the image super-resolution task are reported in Tab.\,\ref{tab:sr}. It can be seen that ERQ consistently outperforms other methods in both the $\times$2 and $\times$4 upscaling factors. 
Specifically, for W4A4 SwinIR$\times$2, ERQ improves PSNR by 0.15 dB, 0.52 dB, 0.56 dB, and 0.54 dB for Set5, Set14, BSD100, and Urban100, respectively. As for W4A4 SwinIR$\times$4, ERQ improves PSNR by 0.30 dB, 0.21 dB, 0.27 dB, and 0.27 dB for Set5, Set14, BSD100, and Urban100, respectively. The qualitative comparison is presented in Fig.\,\ref{fig:sr-results}. We selectively compare the proposed ERQ with PTQ4SR, which is designed for the image super-resolution task, as well as other methods that demonstrate strong quantitative performance. It can be seen that ERQ achieves the best visualization results.
The above results further demonstrate the effectiveness and generalization ability of the proposed ERQ.

\begin{table*}[ht]
\centering
\hspace{-7mm}
% \resizebox{\textwidth}{!}{
\begin{minipage}{.24\textwidth}
\centering
\small
\caption{Ablation studies of $\mathrm{k}$ in Rounding Refinement.}
\begin{tabular}{cc}
\toprule[1.25pt]
 $\mathrm{k}$  &\textbf{Top-1 Acc. (\%)} \\
\midrule[0.75pt]
\midrule[0.75pt]
 0 & 74.06 \\
1  &   \underline{\textbf{74.35}} \\
 2  &  73.99 \\
 3  &  73.79 \\
\bottomrule[1.0pt]
\end{tabular}
\label{tab:abl-k}
\end{minipage}
\hspace{-4mm}
\begin{minipage}{.24\textwidth}
\centering
\small
\caption{Ablation studies of $|\mathcal{O}|$.}
\begin{tabular}{cc}
\toprule[1.25pt]
$|\mathcal{O}|$ (\% of $D_{out}$)     & \textbf{Top-1 Acc. (\%)} \\ 
\midrule[0.75pt]
\midrule[0.75pt]
0  &      72.76              \\ \hline
1 &      74.29              \\ \hline
2 &      74.31             \\ \hline
5 &       \underline{\textbf{74.35}}             \\ \hline
10 &     73.28             \\  \hline
20 &     73.26             \\ 
\bottomrule[1.0pt]
\end{tabular}
\label{tab:size-of-O}
\end{minipage}
\hspace{4.5mm}
\begin{minipage}{.24\textwidth}
\centering
\small
\caption{Ablation studies of maximum iterations \text{T}.}
\begin{tabular}{cc}
\toprule[1.25pt]
\text{T}     & \textbf{Top-1 Acc. (\%)} \\ 
\midrule[0.75pt]
\midrule[0.75pt]
0  &      74.06              \\ \hline
5 &      74.10              \\ \hline
10 &      74.20            \\ \hline
20 &      \underline{\textbf{74.35}}             \\  \hline
50 &      \underline{\textbf{74.35}}             \\ 
\bottomrule[1.0pt]
\end{tabular}
\label{tab:T}
\end{minipage}
\hspace{-5mm}
\begin{minipage}{.24\textwidth}
\centering
\small
\caption{Ablation studies of percentiles in Outlier Channel Selection.}
\begin{tabular}{cc}
\toprule[1.25pt]
 \textbf{Percentile $\tau$}  &  \textbf{Top-1 Acc. (\%)} \\ 
\midrule[0.75pt]
\midrule[0.75pt]
 \{ $\tau^{95}$, $\tau^{5}$ \}  &       73.68              \\ \hline
 \{ $\tau^{98}$, $\tau^{2}$ \}  &       73.95              \\ \hline
 \{ $\tau^{99}$, $\tau^{1}$ \} &      \underline{\textbf{74.35}}             \\  \hline
 \{ $\tau^{99.5}$, $\tau^{0.05}$ \} &       74.20            \\ \hline
 \{ $\tau^{99.9}$, $\tau^{0.01}$ \} &      74.00             \\ 
% 99.99  &     73.99             \\ 
\bottomrule[1.0pt]
\end{tabular}
\label{tab:tau}
\end{minipage}
% }
\end{table*}

\subsection{Ablation Study}

All ablation studies are conducted on the W4A4 DeiT-S.

\subsubsection{Ablation Study of Proposed Components}

In Tab.\,\ref{tab:abl-components}, we report the effect of using Aqer and Wqer. It can be observed that, compared to the baseline, Aqer enhances accuracy by a significant 30.58\% and Wqer also presents a notable 10.54\% performance gain. Then, we analyze the effect of the components within Aqer by setting the baseline as only performing Wqer. As indicated in Tab.\,\ref{tab:abl-Aqer}, Reparameterization Initialization and Ridge Regression respectively provide 22.03\% and 8.67\% accuracy increases, indicating their effectiveness. Combining these two components further presents the highest 23.43\% increase.
Applying all these components provides the best accuracy of 74.35\%. In Tab.\,\ref{tab:abl-Wqer}, we evaluate the effect of the components within Wqer by setting the baseline as only performing Aqer. Compared to the baseline, Dual Uniform Quantization, Rounding Refinement, and Ridge Regression achieve 2.07\%, 0.84\%, and 1.16\% accuracy gains, respectively. The combination of any two of these three components leads to a larger increase. For example, using both Dual Uniform Quantization and Rounding Refinement yields 2.74\% increases. Finally, applying all these components provides the best accuracy of  74.35\%.
These results confirm the effectiveness of the components in ERQ.

\subsubsection{Ablation Study of Hyperparemeters}

\begin{figure}[ht]
\centering
\centerline{\includegraphics[width=0.7\columnwidth]{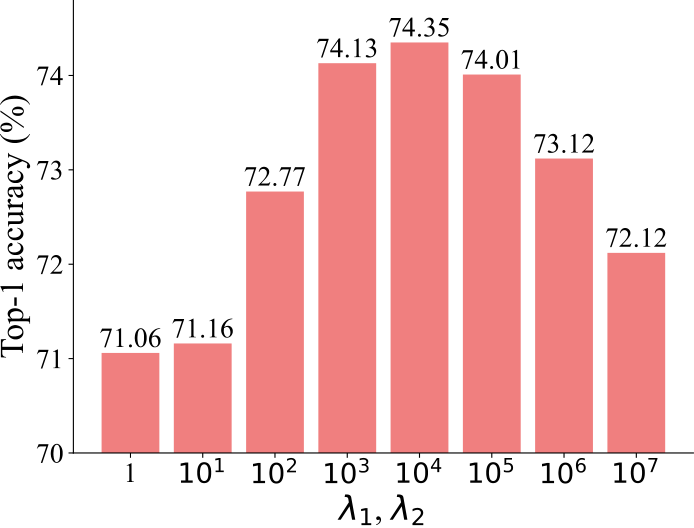}}
\caption{Ablation studies of $\lambda_1$ and $\lambda_2$.}
\label{fig:lambda}
\end{figure}

In Fig.\,\ref{fig:lambda}, we provide the ablation study of $\lambda_1$ of Eq.\,\ref{eq:obj-act1} and $\lambda_2$ of Eq.\,\ref{eq:obj-weight7}. We set $\lambda_1=\lambda_2$ and search for the best value for simplicity. Despite this may not be the best choice, it already yields desirable performance. As shown in Fig.\,\ref{fig:lambda}, the best performance is achieved when $\lambda_1=\lambda_2=1e4$ for W4A4 Deit-S.

In Tab.\,\ref{tab:abl-k}, we present the ablation study of different $\mathrm{k}$ in Eq.\,\ref{eq:obj-weight5} of Rounding Refinement. When $\mathrm{k}=1$, the best accuracy is achieved. Note that when $\mathrm{k}=0$, the Rounding Refinement is invalid.

In Tab.\,\ref{tab:size-of-O}, we present the ablation study of different $|\mathcal{O}|$ in Eq.\,\ref{eq:select} of Dual Uniform Quantization. It can be seen that only separately handling 1\% of $D_{out}$ outlier channel already yields a notable performance gain. When $|\mathcal{O}|$ is set to the 5\% of $D_{out}$, the best accuracy is achieved. Note that when using 0\% of $D_{out}$, the Dual Uniform Quantization is invalid.

Tab.\,\ref{tab:T} presents the ablation study of maximum iterations \text{T} in Alg.\,\ref{alg:wqer}. It can be seen that the best performance is achieved if \text{T}$=20$. After 20, continue increasing \text{T} won't yield gains. Note that when \text{T}$=0$, the Rounding Refinement is invalid.

In Tab.\,\ref{tab:tau}, we present the ablation study of upper percentile and low percentile used in Alg.\,\ref{alg:maximum_outlier_coverage}. Here, we use symmetric percentiles to avoid excessive hyperparameter searching. The results show that the model shows the best accuracy when using $\tau^{99}$, $\tau^{1}$.

\begin{figure}[ht]
\centering
\centerline{\includegraphics[width=0.8\columnwidth]{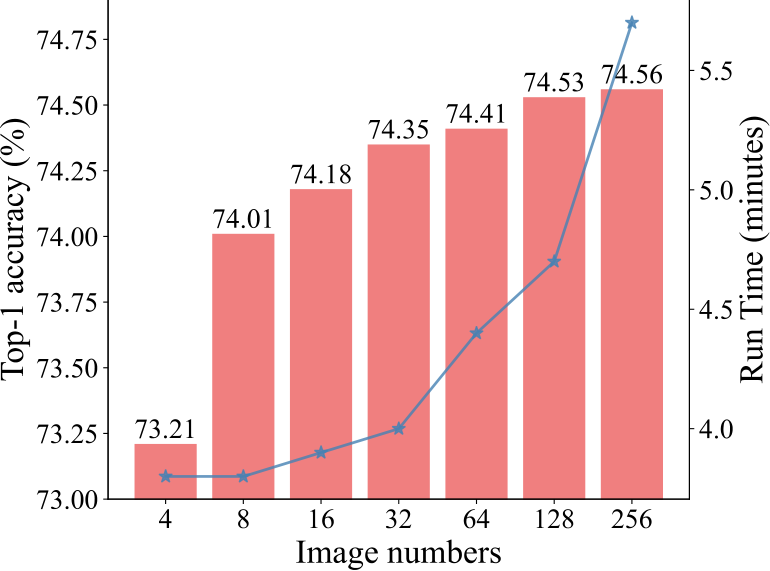}}
\caption{Ablation studies of image number.}
\label{fig:datanumber}
\end{figure}

\subsubsection{Ablation Study of Image Numbers}

Fig.\,\ref{fig:datanumber} presents the ablation study using different image numbers. We further provide the time costs for different numbers of images.
As the image number increases, the performance and time costs increases correspondingly. For example, the accuracy is  73.21\% and 74.01\% for 4 and 8 images, respectively. Then, the performance reaches the plateau after 128 images. Despite more images improving the performance, we adopt 32 images to align with the previous study~\cite{li2023repq} for a fair comparison. 

\begin{table}[ht]
  \centering
  \small
  \caption{Ablations studies of the order of Aqer and Wqer. ``Wqer$\longrightarrow$Aqer'' and ``Aqer$\longrightarrow$Wqer'' respectively represent executing Wqer and Aqer first.}
  \begin{tabular}{cc}
  \toprule[1.25pt]
    \textbf{Error Reduction Order} &  \textbf{Top-1 Acc. (\%)} \\ \midrule[0.75pt]
  \midrule[0.75pt]
     Wqer$\longrightarrow$Aqer       & 73.05  \\            
    Aqer$\longrightarrow$Wqer   &  \underline{\textbf{74.35}}  \\\bottomrule[1.0pt]
  \end{tabular}
  \label{tab:abl-order}
\end{table}

\begin{table}[!ht]
\centering
\small
\caption{{ Time costs and the number of optimizable parameters (in brackets) of different methods. ``*'' indicates the results are re-produced by using the official code.}}
% \resizebox{0.45\textwidth}{!}{%
\begin{tabular}{cccc}
\toprule[1.25pt]
\textbf{Method} &  \textbf{Runtime}     & \textbf{Top-1 Acc. (\%)} \\ 
\midrule[0.75pt]
\midrule[0.75pt]
BRECQ*~\cite{li2021brecq}            & $\sim$48 minutes { (22.04M)}               & 32.89              \\ \hline
QDrop*~\cite{wei2021qdrop}           & $\sim$80 minutes { (22.04M) }      & 35.79             \\ \hline
PD-Quant*~\cite{liu2023pd}              & $\sim$110 minutes { (22.04M)}           & 64.85             \\ \hline
OAS-ViT~\cite{maoutlier}          & $\sim$50 minute { (22.04M)  }              & 37.15             \\ \hline
GPTQ*~\cite{frantar-gptq}            & $\sim$3 minutes { (22M)}                & 70.85              \\ \hline
RepQ-ViT~\cite{li2023repq}          & $\sim$1 minute { (-) }               & 69.03             \\ \hline
ERQ (Ours)           & $\sim$4 minutes  { (22M)}               &   \underline{\textbf{74.35}}            \\  \bottomrule[1.0pt]
\end{tabular}
% }
\label{tab:timecosts}
\end{table}

\subsubsection{Ablation Study of the Order of Aqer and Wqer}

Tab.\,\ref{tab:abl-order} presents the ablation study using different orders of Aqer and Wqer. 
It can be seen that exchanging the order of Aqer and Wqer leads to performance degradation since the quantized weights cannot well accommodate the $\delta\mathbf{W}^*$ (Eq.\,\ref{eq:eq10}).

\begin{table}[!ht]
  \centering
  \small
  \caption{Ablation studies of the partition strategy in Wqer. "Smallest," "Largest," and "Half" represent the strategies of quantizing FP weights with the smallest quantization error first, the largest quantization error first, and the first half of FP weights, respectively.}
  \begin{tabular}{ccc}
  \toprule[1.25pt]
  \textbf{Partition Strategy} &  \textbf{Runtime}&  \textbf{Top-1 Acc. (\%)} \\ \midrule[0.75pt]
  \midrule[0.75pt]
      Smallest  &  $\sim$20 minutes &   74.19 \\   
     Largest   &  $\sim$20 minutes  &   73.90 \\
      Half  &  $\sim$4 minutes &   \underline{\textbf{74.35}}\\\bottomrule[1.0pt]
  \end{tabular}
  \label{tab:abl-partition}
\end{table}

\subsubsection{Ablation Study of the Partition Strategy}

In Wqer, the FP weights are divided into two segments: the first half and the second half, with the first half quantized first. This strategy ensures a fixed order for quantizing the weights across all channels, enabling efficient parallel processing.
Here, we conduct an ablation study on two partition strategies: quantizing FP weights starting with the smallest quantization error first and starting with the largest quantization error first. The results in Tab.\,\ref{tab:abl-partition} indicate that these two strategies typically hurt performance. Furthermore, using these strategies alters the order of weight quantization for each output channel, rendering parallel processing infeasible and resulting in higher time costs compared to the current partition strategy. For example, time costs increase from 4 minutes with the current strategy to 20 minutes when using the smallest or largest error first.

\subsection{Comparisons of Time Costs}

Tab.\,\ref{tab:timecosts} compares the time costs and the number of optimizable parameters between the proposed ERQ and other PTQ methods. 
Regarding time costs, BRECQ, QDrop, and PD-Quant require longer time overhead. In contrast, GPTQ, RepQ-ViT, and the proposed ERQ demonstrated significantly reduced time costs. Notably, our ERQ achieved the best Top-1 Accuracy of 74.35\% with a runtime of only 4 minutes. In terms of the number of optimizable parameters, ERQ maintains a comparable count to most other methods, except for RepQ-ViT, which performs only calibration and thus has no optimizable parameters. Importantly, whether compared with iterative update methods such as BRECQ and QDrop or calibration-based methods like RepQ-ViT, ERQ achieves significantly better performance with comparable or even reduced running time.

\begin{figure}[ht]
\centering
\centerline{\includegraphics[width=\columnwidth]{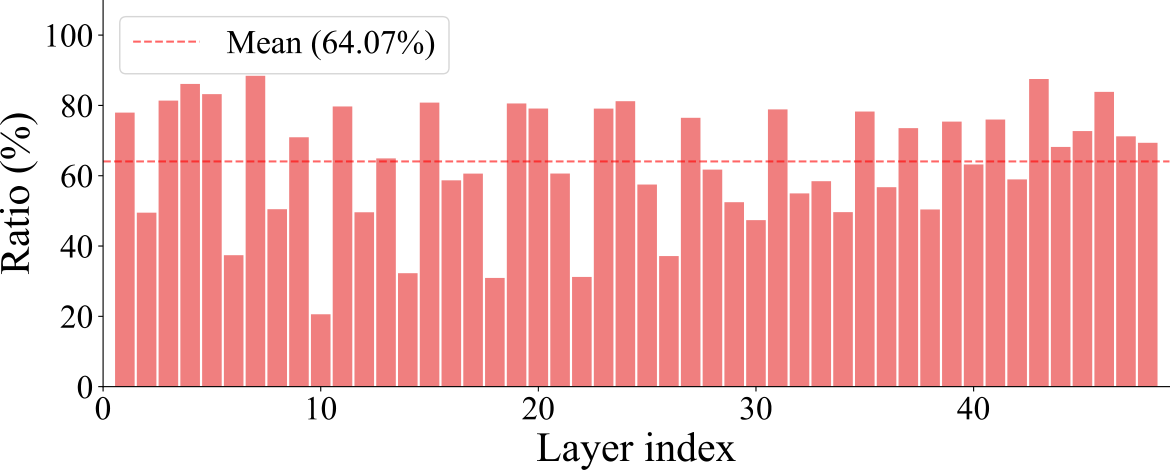}}
\caption{Illustration of the MSE (evaluated by Eq.\,\ref{eq:obj}) reduction ratio brought by ERQ. We plot the error reduction ratio for each layer.}
\label{fig:error-reduction}
\end{figure}

\subsection{Comparisons of Real Inference Runtime}

\begin{table}[!ht]
\centering
\small
\caption{Comparisons of real inference time. The number in the bracket is the speedup compared with full-precision. Input batch size: 200.}
\begin{tabular}{ccc}
\toprule[1.25pt]
 \diagbox{\textbf{Bit-width}}{\textbf{Model}}  &  \textbf{ViT-B} &  \textbf{ViT-L}  \\ 
\midrule[0.75pt] 
\midrule[0.75pt]
 Full-Precision   &  524.2ms & 1577.1ms\\
 8-bit   &   362.6ms (1.38x) &  814.1ms (1.94x) \\
 4-bit   &   244.9ms (2.14x) &  515.5ms (3.06x) \\
 DUQ 4-bit   &   256.8ms (2.04x) & 
  517.9ms (3.06x) \\
\bottomrule[1.0pt]
\label{tab:real-time}
\end{tabular}
\end{table}

To demonstrate the efficiency of DUQ, we implemented CUDA kernels for quantization, matrix multiplication, and dequantization based on the Cutlass library\footnote{https://github.com/NVIDIA/cutlass}. As shown in Tab.~\ref{tab:real-time}, we conducted experiments on ViT-B and ViT-L. Note that for LN and softmax, we still use high-precision computations.
The benchmarks are evaluated on an NVIDIA 3090 GPU, with speedup measured by averaging the results of 10 runs. We observed that DUQ 4-bit quantization incurs negligible additional inference time compared to standard 4-bit quantization, for example, 517.9ms versus 515.5ms for ViT-L. Furthermore, 4-bit DUQ quantization achieves significant speedup over full-precision. For example, it achieves 3.06x speedup over the full-precision on ViT-L.

\subsection{Validation of Error Reduction}

We plot the MSE reduction ratio brought by our ERQ in Fig.\,\ref{fig:error-reduction}. The reduction ratio is computed using the value of Eq.\,\ref{eq:obj} on W4A4 DeiT-S. It can be observed that the proposed ERQ effectively reduces the MSE by achieving a 20\% - 90\% reduction ratio. Notably, ERQ yields a significant 64.07\% average MSE reduction, supporting the performance results as shown in Tab.\,\ref{tab:imagenet-0}.

For a comprehensive comparison, we also evaluate the error reduction rate achieved by the joint optimization of Eq.\,\ref{eq:obj} with iterated, gradient-based SGD and Adam optimizers.
The optimization ran for 10,000 iterations with an initial learning rate of 1e-6 and a cosine learning rate schedule. The error reduction rates for SGD and Adam are 33.18\% and 35.72\%, respectively, both of which are lower than the proposed ERQ, which achieved an average reduction of 64.07\% as shown in Fig.,\ref{fig:error-reduction}. Accordingly, using SGD and Adam, the top-1 accuracy on ImageNet reached 73.23\% and 73.59\%, respectively, while ERQ achieved 74.35\%. Furthermore, SGD and Adam respectively took 35 and 39 minutes, while ERQ required only about 4 minutes. This demonstrates that ERQ not only outperforms in terms of accuracy but also significantly reduces computational time. In summary,
these findings justify the performance improvements achieved by ERQ, which addresses activation and weight quantization errors sequentially.

\section{Conclusion}

In this paper, we present ERQ, a two-step PTQ method of ViTs, consisting of Activation quantization error reduction (Aqer) and Weight quantization error reduction (Wqer) to respectively mitigate the quantization error induced by activation and weight quantization. 
Aqer first employs Reparameterization Initialization for high-variance post-LayerNorm activations to mitigate the initial activation quantization errors. Then, it formulates a Ridge Regression problem to further tackle quantization errors by updating full-precision weights with a closed-form solution.
Wqer first incorporates Dual Uniform Quantization for outlier-extensive weights to mitigate the initial weight quantization errors. Then, it progressively tackles the error in a quantization-and-correction manner.
At each iteration, the first half of weights are quantized and the resulting error is first mitigated by Rounding Refinement and then again by solving a Ridge Regression problem.
The former mitigates the errors by leveraging an empirically derived efficient proxy of output error to refine the rounding directions of quantized weights.
The latter further tackles the errors by again formulating a Ridge Regression to update the remaining full-precision weights with a closed-form solution.
The effectiveness of ERQ is demonstrated by extensive experiments on various ViTs variants across diverse tasks.

\ifCLASSOPTIONcompsoc
  \section*{Acknowledgments}
\else
  \section*{Acknowledgment}
\fi

This work was supported by National Science and Technology Major Project (No. 2022ZD0118202), the National Science Fund for Distinguished Young Scholars (No.62025603), the National Natural Science Foundation of China (Np. 624B2119, No. U21B2037, No. U22B2051, No. 62176222, No. 62176223, No. 62176226, No. 62072386, No. 62072387, No. 62072389, No. 62002305 and No. 62272401), and the Natural Science Foundation of Fujian Province of China (No.2021J01002,  No.2022J06001).

\ifCLASSOPTIONcaptionsoff
  \newpage
\fi



%\bibliographystyle{IEEEtran}
\bibliographystyle{IEEEtran}
\bibliography{example_paper}

\end{document}